\documentclass[twoside]{article}

\usepackage[preprint]{aistats2026}

\usepackage{shortex} 
\usepackage[utf8]{inputenc} 
\usepackage[T1]{fontenc}    
\usepackage{hyperref}       
\usepackage{url}            
\usepackage{booktabs}       
\usepackage{amsfonts}       
\usepackage{nicefrac}       
\usepackage{microtype}      
\usepackage{xcolor}         
\usepackage{algorithm}
\usepackage{algpseudocode}
\usepackage{pifont}
\usepackage{accents}
\usepackage{enumitem}

\definecolor{julia1}{rgb}{0, 0.60, 0.98}
\definecolor{julia1dark}{rgb}{0, 0.20, 0.68}
\definecolor{julia2}{rgb}{0.89, 0.43, 0.28}
\definecolor{julia3}{rgb}{0.24, 0.64, 0.30}
\definecolor{julia4}{rgb}{0.76, 0.44, 0.82}
\definecolor{julia5}{rgb}{0.67, 0.55, 0.09}
\definecolor{julia6}{HTML}{00aaae}
\definecolor{julia7}{HTML}{ed5e93}
\definecolor{julia8}{HTML}{c68125}
\definecolor{julia9}{HTML}{00a98d}

\newcommand{\anonymize}[1]{}
\newcommand{\anonymizewithtext}[1]{\textit{\textcolor{gray}{(removed to preserve anonymity)}}}

\newcounter{xxx}
\begin{document}

\twocolumn[

\aistatstitle{AutoGD: Automatic Learning Rate Selection for Gradient Descent}

\aistatsauthor{Nikola Surjanovic \And Alexandre Bouchard-C\^{o}t\'{e} \And Trevor Campbell }

\aistatsaddress{ University of British Columbia \And University of British Columbia \And University of British Columbia } ]

\begin{abstract}
  The performance of gradient-based optimization methods, 
  such as standard gradient descent (GD), 
  greatly depends on the choice of learning rate. 
  However, it can require a non-trivial amount of user tuning effort 
  to select an appropriate learning rate schedule.
  When such methods appear as inner loops of other algorithms, 
  expecting the user to tune the learning rates may be impractical.
  To address this, we introduce AutoGD: a gradient descent method that 
  automatically determines whether to increase or decrease the learning 
  rate at a given iteration.  
  We establish the convergence of AutoGD, and show that we can 
  recover the optimal rate of GD (up to a constant) for a broad class of functions 
  without knowledge of smoothness constants.
  Experiments on a variety of traditional problems and variational inference 
  optimization tasks demonstrate strong performance 
  of the method, along with its extensions to AutoBFGS and AutoLBFGS.


\end{abstract}
\section{INTRODUCTION}
\label{sec:intro}

Gradient descent (GD) and its many popular variants---such as 
backtracking line search and higher-order methods 
\cite{nocedal2006numerical,armijo1966minimization}---are 
indispensable tools for solving optimization problems with moderate dataset sizes. 
For instance, such optimization problems are frequently encountered in statistics 
in the context of maximum likelihood estimation of model parameters. 
For black-box variational inference tasks, deterministic first- and second-order 
methods are suitable when sample average approximation (SAA) techniques 
are used to approximate the objective \cite{giordano2024dadvi,burroni2023saa}.
Further, for any optimization task where the objective function cannot be expressed 
as a sum of terms, noiseless gradient descent methods serve as 
extremely useful optimization tools.

\begin{figure*}[t]
  \centering
  \begin{subfigure}{0.32\textwidth}
    \centering
    \includegraphics[width=\textwidth]{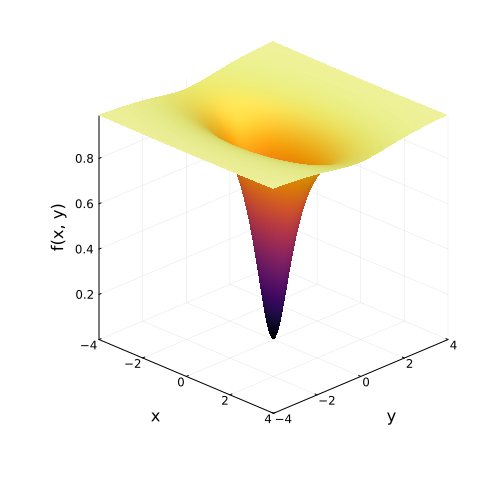}
  \end{subfigure}
  \begin{subfigure}{0.32\textwidth}
    \centering
    \includegraphics[width=\textwidth]{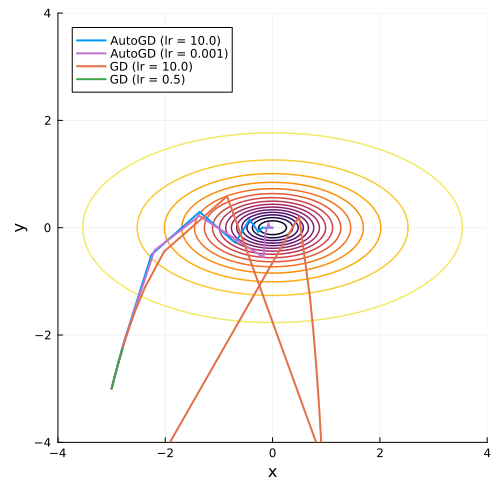}
  \end{subfigure}
  \begin{subfigure}{0.32\textwidth}
    \centering
    \includegraphics[width=\textwidth]{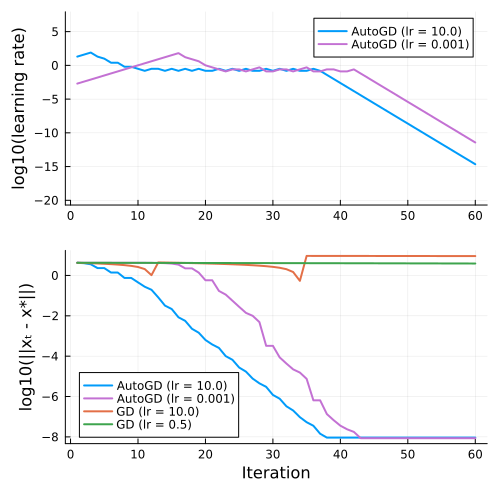}
  \end{subfigure}
  \caption{Performance of deterministic optimizers on the 
  non-convex objective function $f(x, y) = 1 - 1/(1 + x^2 + 4y^2)$. 
  \textbf{Left:} Surface plot of the objective function. 
  \textbf{Middle:} Trajectories of AutoGD with initial learning rates $\gamma_0 \in \{0.001, 10.0\}$ 
  and GD with learning rates $\gamma \in \{0.5, 10.0\}$ over 100 iterations. 
  Here, GD with $\gamma = 0.5$ converges very slowly, while $\gamma = 10.0$ is unstable.
  AutoGD is stable as it approaches the minimum for different initial learning rate values.
  \textbf{Top right:} Automatically selected learning rates (on log scale) for each of the first 60 iterations.
  AutoGD automatically learns to anneal the learning rate in the initial phase, and then 
  decreases the learning rate upon convergence. 
  \textbf{Bottom right:} Distance to optimum (log scale) for AutoGD and GD iterates.}
  \label{fig:autogd}
\end{figure*}

An important goal of modern optimization algorithms is to provide efficient convergence 
to an optimum of the objective function with as little tuning effort required by the user as possible. 
Often such optimization tasks appear as inner loops of other algorithms, 
hidden away from the user, and so the robustness of the underlying optimizers is of utmost importance.
For instance, this may be the case for some penalized likelihood methods, 
where regularization parameters are updated via cross-validation in an outer loop while other parameters 
are tuned with gradient descent \cite{hastie2009elements}. 
Another example is the common expectation--maximization (EM) algorithm 
with a non-closed-form maximization update, requiring the use of a first- or second-order 
inner loop update inside the EM outer loop \cite{ruth2024em}. 

Gradient descent algorithms each differ in their set of tuning parameters, but the 
learning rate (sequence) is common among almost all such methods and is critical to performance.  
If the learning rate is chosen to be too large, GD may become unstable or diverge; whereas 
if the learning rate is too small, the iterates may converge but at a painstakingly 
slow pace. Further, there is usually no ``one size fits all'' learning rate for a given 
optimization problem as different regions of the parameter space during optimization may benefit 
from larger or smaller learning rates due to varying curvature. 

In this work we introduce AutoGD, a new GD algorithm that 
adaptively selects appropriate learning rates on the fly by comparing the 
performance of neighbouring (larger and smaller) learning rates (see \cref{fig:autogd}). 
A preliminary version of AutoGD was introduced in \cite{surjanovic2025autosgd} 
and studied only briefly.
Here we establish that AutoGD converges
under appropriate assumptions on the objective function, providing both 
asymptotic results and nonasymptotic bounds. 
These rates of convergence are equivalent (up to a constant) to the optimal 
rate of convergence of gradient descent on $L$-smooth and $\mu$-strongly convex functions,
but crucially do not require knowledge of either parameter.
In our experiments we find that AutoGD is extremely robust and outperforms other 
gradient descent methods or
is comparable to first-order methods that also require no tuning effort.
We extend the methodology to second-order methods such as BFGS and L-BFGS, 
resulting in the methods AutoBFGS and AutoLBFGS.
We verify the performance of such methods empirically but leave the theoretical analysis 
of second-order methods for future work.

\noindent \textbf{Related work.} 
Among the more traditional approaches to selecting learning rates for gradient descent, 
Polyak step sizes \cite{polyak1969minimization,barre2020polyak} and line search methods 
\cite{armijo1966minimization,wolfe1969convergence,truong2021backtracking}
have been studied extensively. Polyak step sizes typically require knowledge 
of the minimum value attained by the 
objective function, which in general cannot be known. 
Line search methods with backtracking and descent/curvature (Wolfe) conditions 
are standard practical approaches to estimating an appropriate learning rate at 
a given gradient descent iteration, and numerous variations have emerged 
\cite{more1994linesearch,grippo1986nonmonotone,zhang2004nonmonotone}.
A drawback of many of these methods is that they may require many 
function evaluations at each iteration.
Finally, Barzilai--Borwein methods \cite{barzilai1988twopoint,raydan1997barzilai,zhou2025adabb}
use the past two iterates to inform the choice of learning rates to approximately satisfy 
secant equations. However, others have noted that 
some of these methods ``lack consistency and may even lead to divergence, 
even for simple convex problems'' \cite{malitsky2024adaptive}.

Recently, hyperparameter-free methods similar in spirit to our proposed AutoGD 
algorithm have been studied in both the deterministic and stochastic settings 
\cite{defazio2023dadaptation,ivgi2023dog,khaled2023dowg,orabona2017coin,kreisler2024parameterfree,mishchenko2023prodigy,malitsky2019adgd,khaled2024tuningfree,loizou2021stochasticpolyak,orvieto2022polyak,attia2024parameterfree}.
We focus in this work on the deterministic optimizers,
such as AdGD and AdGD2 \cite{malitsky2019adgd,malitsky2024adaptive},
and demonstrate several scenarios where AdGD methods can fail to converge 
or struggle. In contrast, for AutoGD we establish both empirical and theoretical robustness, 
recommending its use as a fully general-purpose black-box optimization algorithm.

Finally, past work has established the convergence of gradient descent and first-order methods to 
a local minimum (or avoidance of a saddle point) under the assumption of a diffuse starting point and a sufficiently 
small learning rate \cite{lee2016gradient,panageas2016gradient,lee2019gradient}. 
In our work we establish that AutoGD converges to a local minimum under very mild conditions 
provided that both the starting point and initial learning rate are diffuse. 
We also use a notion of an unstable saddle (\cref{defn:unstable}), which 
allows for higher-order saddles with non-negative eigenvalues.

\section{SETUP} 
\label{sec:background}

The goal is to find a minimum of a differentiable objective function $f: \reals^d \to \reals$. 
Without loss of generality, we assume that $f$ is nonnegative 
with $\inf_x f(x) = 0$, as none of our proposed methods require knowledge of the 
minimum value attained by $f$.

Starting with an initial iterate $x_0 \in \reals^d$ and
a sequence of learning rates $(\gamma_t)_{t \geq 0}$,
the standard GD algorithm is defined by the sequence 
$(x_t)_{t \geq 0}$ produced by 
\[
\label{eq:GD}
  x_{t+1} = x_t - \gamma_t \nabla f(x_t). 
\]
In what follows, we address how to automatically choose the learning rate sequence 
$(\gamma_t)_{t \geq 0}$ with minimal knowledge of the objective function. 
\section{AUTOGD} 
\label{sec:autogd}

The AutoGD method has the following inputs: 
an initial iterate $x_0$, 
initial baseline learning rate $\gamma_0$, 
learning rate scaling factor $c > 1$,
and Armijo constant $0 < \eta < (c+1)/(c^2+1)$.
At a given iteration, the AutoGD algorithm assesses several learning rates 
$\{c^{-1} \gamma, \gamma, c\gamma\}$ centered around the current baseline $\gamma$. 
After each iteration, the baseline learning rate can either increase, decrease, or stay the same. 

We initialize $x_0,\gamma_0$ randomly (e.g., $x_0 \sim \Norm(\mu_0, \sigma^2 I)$ and 
$\log \gamma_0 \sim \Norm(0, \sigma^2)$ for some very small $\sigma^2 > 0$). 
The only condition on the initialization we require is that $x_0, \gamma_0$ are 
initialized from a distribution dominated by the Lebesgue measure on $\reals^d \times \reals_{>0}$, 
but can otherwise be arbitrary.
Then for each $t\in\{0,1,\dots\}$, we define the proposed new learning rate
\[
  \gamma'_{t+1} &= \argmin_{\gamma \in \{0, c^{-1}\gamma_t,\gamma_t,c\gamma_t\}}\quad 
    f(x_t - \gamma\grad f(x_t)) \\
  &\text{s.t. } \, \, f(x_t - \gamma\grad f(x_t)) \leq f(x_t) - \eta\gamma\|\grad f(x_t)\|^2.
\]
The ability to choose $\gamma = 0$ in the optimization guarantees that the
feasible set is nonempty, and the choice $\gamma=0$ represents a ``no
movement'' option in the event that all learning rate choices result in
insufficient objective decrease.  If multiple feasible learning rates result in the
same objective value, we select the smallest; this ensures that if $x_{t+1} \neq
x_t$, then $f(x_{t+1}) < f(x_t)$.  
We then update the state
\[
  x_{t+1} \gets x_t - \gamma_{t+1}' \grad f(x_t)
\]
using the ``lookahead'' learning rate $\gamma_{t+1}'$.
The next baseline learning rate is chosen to be
\[
  \gamma_{t+1} = 
  \begin{cases}
    \gamma_t, & \|\grad f(x_t)\|= 0, \\ 
    \gamma'_{t+1}, & \|\grad f(x_t)\| \neq 0, \, \gamma'_{t+1} \neq 0, \\
    c^{-2}\cdot\gamma_t, & \|\grad f(x_t)\| \neq 0, \, \gamma'_{t+1} = 0.
  \end{cases}
\]
Note that if $\gamma_{t+1}' = 0$, we ensure that we make no movement and also 
decrease the baseline learning rate for the next iteration.
We shrink by $c^{-2}$ in this case because we already know that both $c^{-1}\gamma_t$ and $\gamma_t$
did not result in a feasible learning rate, ensuring that we do
not spend time re-examining these learning rates.
Further, keeping $\gamma_t$ constant when $\|\grad f(x_t)\|= 0$ does not matter in
practice (since the algorithm has already converged to a stationary point), 
but simplifies the theoretical analysis.

\begin{algorithm}[t]
	\begin{algorithmic}[1]
    \Require State and learning rate $(x_t, \gamma_t)$
    \Require Scaling coefficient $c > 1$ (default: $c = 2$)
    \Require Armijo constant $0 < \eta < (c+1)/(c^2+1)$ (default: $\eta = 10^{-4}$).
      \State $f_0 \gets f(x_t)$ 
      \State $g \gets \nabla f(x_t)$
      \State $\texttt{valid} \gets \{0\}$ \Comment{Always include 0 in \texttt{valid}}
      \LineComment{Parallelize the following loop}
      \For{$\gamma$ {\bf in} $\{c^{-1} \gamma_t, \gamma_t, c \gamma_t\}$}
        \State $f_\gamma \gets f(x_t - \gamma g)$
        \If{$f_\gamma \leq f_0 - \eta\gamma\|g\|^2$} 
          \State $\texttt{valid} \gets \texttt{valid} \cup \{\gamma\}$ 
        \EndIf
      \EndFor
      \LineComment{In case of ties, choose the smallest $\gamma$}
      \State $\gamma'_{t+1} \gets \argmin_{\gamma \in \texttt{valid}} f_\gamma$ 
      \State $x_{t+1} \gets x_t - \gamma_{t+1}' g$
      \If{$\|g\| = 0$}
        \State $\gamma_{t+1} \gets \gamma_t$ 
      \ElsIf{$\gamma'_{t+1} = 0$}
        \State $\gamma_{t+1} \gets c^{-2} \gamma_t$ 
      \Else 
        \State $\gamma_{t+1} \gets \gamma_{t+1}'$ 
      \EndIf
    \State \Return $(x_{t+1}, \gamma_{t+1})$
	\end{algorithmic}
  \caption{One Step of AutoGD}
  \label{alg:autogd}
\end{algorithm}

The complete AutoGD algorithm is presented in \cref{alg:autogd}.
\cref{fig:autogd} shows the performance of AutoGD on a non-convex objective function 
and how the chosen learning rate varies as the algorithm proceeds.
In terms of the computational complexity of AutoGD compared to standard 
gradient descent, the main difference between the two algorithms is that the former 
evaluates three learning rates at each iteration instead of one. 
However, because these evaluations can be performed in parallel and independently from one 
another, an efficient implementation of AutoGD can have the same average runtime as GD.

We make some comments with respect to the tuning parameters of AutoGD. 
Although AutoGD works with a wide range of reasonable values of $c$, we recommend 
a default choice of $c = 2$, which corresponds to doubling/halving the learning rate. 
Also, even if $\gamma_0$ is initially too small or too large, we reach an 
appropriate learning rate $\gamma$ within $\abs{\log_c(\gamma/\gamma_0)}$ iterations,
and so $\gamma_0$ can be set anywhere within a wide acceptable neighbourhood.
We suggest to initialize with $\log \gamma_0 \sim \Norm(0, 10^{-12})$.
In practice, $\eta$ should be chosen to be very small, e.g.~$\eta \approx 10^{-4}$.

We now justify various components of the algorithm: the ``no movement'' 
option ($\gamma = 0$), the Armijo rule, and the diffuse initialization. 
Counterexamples demonstrate failure modes if these components are omitted.

\subsection{Importance of the ``no movement'' option} 
\label{sec:no_movement} 

The following result highlights why it is important to consider a 
``no movement'' step in the algorithm, where we set $\gamma_{t+1}' = 0$ and 
decrease the learning rate by $c^{-2}$.
Suppose we were to optimize over $\{c^{-1}\gamma, \gamma, c\gamma\}$
instead of $\{0, c^{-1}\gamma, \gamma, c\gamma\}$ at each iteration (and without use
of the Armijo condition check), even if an objective function increase is detected 
at all $\gamma$ values.
The algorithm would then diverge even on some simple polynomials.
However, introducing the ``no movement'' option resolves this case, 
as we will see in \cref{sec:theory}.
The following results assumes that we must select a learning rate in $\{c^{-1} \gamma, \gamma, c\gamma\}$ 
and set that as the baseline learning rate for the next iteration. 

\bcexa 
\label{result:autogd_diverges}
Consider iterates $(x_t)_{t \geq 0}$ of AutoGD with initial learning rate 
$\gamma_0$ and where at each iteration we update our iterates without 
the ``no movement'' option.
Suppose $\abs{x_0} > 1$ and that $f(x) = x^{2p}$ for $p \in \nats$, $p \geq 2$ with 
\[
  p > \frac{c(c+1)}{2 \gamma_0\abs{x_0}}.
\]
Then the iterates diverge exponentially: $\abs{x_t} = \Omega(c^t)$.
\ecexa

\subsection{Importance of the Armijo constant} 
\label{sec:armijo}

\begin{figure*}[!t]
  \centering
  \begin{subfigure}{0.4\textwidth}
    \centering
    \includegraphics[width=\textwidth]{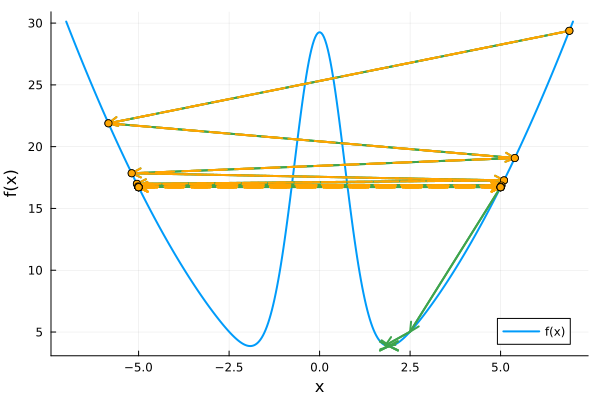}
  \end{subfigure}
  \begin{subfigure}{0.4\textwidth}
    \centering
    \includegraphics[width=\textwidth]{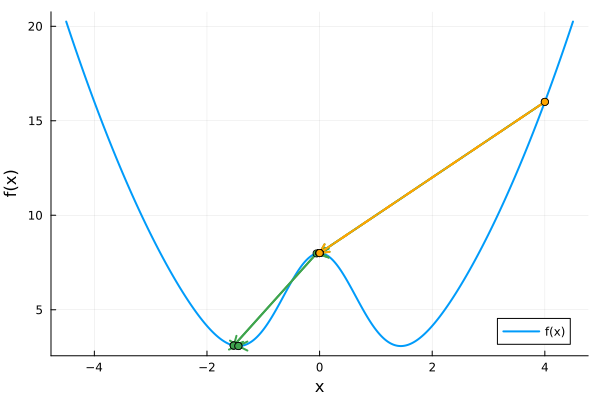}
  \end{subfigure}
  \caption{Two counterexamples demonstrating the importance of the Armijo 
  condition and diffuse initialization for AutoGD. 
  \textbf{Left:} \cref{prop:counterexample_1}. Orange dashed arrows 
  indicate AutoGD without the Armijo condition converging to a cycle. In contrast, 
  AutoGD with the Armijo condition (green) converges to a local minimum. 
  \textbf{Right:} \cref{prop:counterexample_2}. 
  Orange dashed arrows indicate AutoGD with a deterministic starting point converging 
  to a local maximum. By using a diffuse initialization (green), AutoGD is 
  able to avoid the local maximum almost surely and converge to a local minimum.}
  \label{fig:counterexamples}
\end{figure*}

There exist nonconvex functions and initializations
$x_0,\gamma_0$ for which the iterates $x_t$ of AutoGD with $\eta=0$ 
can converge to a limit cycle of strictly decreasing $f(x_t)$ with $\grad f(x_t)
\not\to 0$. Setting a small $\eta > 0$ ensures that this does not occur.
In practice, the Armijo constant $\eta$ should be chosen to be very small, (e.g., $\eta =
10^{-4}$). 
An illustration of this counterexample is given in the left panel of 
\cref{fig:counterexamples}, along with a demonstration of the performance of 
AutoGD with $\eta > 0$. 

\bcexa
\label{prop:counterexample_1}
Consider the function $f:\reals \to\reals$ defined by
\[
  f(x) = |x|^{7/4} + b\exp\lt(-x^2\rt), \qquad b >0.
\]
Fix constants $\stx,\delta > 0$ and set
\[
  b &= \frac{7\stx^{7/4}}{4(1-\exp(-\stx^2))}, \qquad
  \gamma_0 = \frac{\stx^{1/4}}{\frac{7}{8}-b\stx^{1/4}\exp\lt(-\stx^2\rt)},
\]
with $x_0 = \stx + \delta$.
Then, for all sufficiently large $\stx$ and sufficiently small $\delta$,
we have that $b > 0$,  $\gamma_0 > 0$, and
the iterations of AutoGD with $c=2$, $\eta = 0$
initialized at $x_0,\gamma_0$ satisfy
\[
\lim_{t\to\infty} x_{2t} = \stx \qquad \lim_{t\to\infty} x_{2t+1} = -\stx.
\]
Therefore, $\grad f(x_t) \not\to 0$ and the iterates do not converge.
\ecexa

\subsection{Importance of the diffuse initialization} 
\label{sec:diffusivity} 

Finally, we demonstrate the importance of initializing $(x_0, \gamma_0)$ from 
a diffuse measure. 
Even with the ``no movement'' option and $\eta > 0$, there may exist initializations 
for which GD and AutoGD converge to a local maximum or a saddle point instead of a 
local minimum as desired. 
This example and its resolution using diffuse initializations are demonstrated 
in \cref{fig:counterexamples}, with supporting theory in \cref{sec:theory}.

\bcexa
\label{prop:counterexample_2}
Consider the function $f:\reals \to\reals$ defined by
\[
  f(x) = x^2 + b\exp\lt(-x^2\rt), \quad b > 0.
\]
Suppose $\eta < 1/4$, $x_0>0$ is large enough such that
\[
  b\lt(1-(1+4\eta x_0^2)\exp\lt(-x_0^2\rt)\rt) &< (1-4\eta)x_0^2 \\ 
  x_0\lt(1-b\exp\lt(-x_0^2\rt)\rt) &> 0,
\]
and set $\gamma_0 = 2x_0/\grad f(x_0)$.
Then, the iterates of AutoGD with $c=2$ initialized at $x_0,\gamma_0$ (i.e., with a non-diffuse initialization) 
converge to the local maximum $x_t \to 0$ as $t\to\infty$.
\ecexa

\section{THEORY} 
\label{sec:theory} 

In this section, we prove both asymptotic and nonasymptotic properties of AutoGD.
Standard definitions, such as (strong) convexity and $L$-smoothness, are deferred to 
the supplementary material.
In what follows, we always assume that $f$ is differentiable and bounded below (with 
$\inf_x f(x) = 0$, without loss of generality).
AutoGD is always used with the ``no movement'' option.
We also always use AutoGD with the Armijo condition and diffuse initialization.

\subsection{Asymptotics} 
\label{sec:asymptotics}

We begin by noting that the function evaluations of iterates produced by AutoGD must always converge.  
This means that AutoGD will never create unstable or divergent iterations. 

\bprop
\label{prop:monotonef}
There exists a $\slf \geq 0$ such that
the iterates $x_t$ of AutoGD satisfy $f(x_t) \downarrow \slf$.
\eprop

With the addition of $L$-smoothness, we can 
conclude that AutoGD asymptotically clusters around a (set of) critical point(s). 

\bthm
\label{thm:tostationarity}
Let $f$ be $L$-smooth. 
The iterates $x_t$ of AutoGD satisfy $\grad f(x_t) \to 0$ as $t\to\infty$.
\ethm

As seen previously, without a diffuse initialization, AutoGD may converge to a 
critical point other than a local minimum or may not converge at all 
(even if $\nabla f(x_t) \to 0$). 
We guarantee that the iterates of AutoGD do not converge to an unstable
point, and no subsequence of iterates converges to a local maximum. 

For the definition of local maxima, we use a definition that is more general
than strict local maxima, but does not include all local maxima.
While usual local maxima can include flat regions, we restrict any flatness
to a zero measure set. 

\bdefn{\emph{(Almost strict local maximum)}}
\label{defn:localmax}
An \emph{almost strict local maximum} is a critical point $x^\star\in\reals^d$ such that there exists 
a neighbourhood
$x^\star \in U$ and set of Lebesgue measure zero $N$ such that
\[
\forall x \in U\setminus N,\quad f(x) < f(x^\star).
\]
\edefn

We use the following definition of an unstable saddle, which allows for higher-order 
saddle points. 

\bdefn{\emph{(Unstable saddle)}}
\label{defn:unstable}
An \emph{unstable saddle} is a critical point $x^\star\in\reals^d$ that is not 
a local maximum, with the property that 
there exists a neighbourhood $x^\star\in U$, direction $v\in\reals^d$, $\|v\|=1$, 
and set of Lebesgue measure zero $N$ such that 
\[
  \forall x \in U\setminus N, \quad (x^\star-x)^Tv\cdot v^T\grad f(x) > 0.
\]
\edefn

The following result shows that AutoGD naturally avoids various undesirable critical points.
For this result we make use of the diffuse initialization to establish almost sure 
avoidance of such points. 

\bthm
\label{lem:avoidbadcritical}
Let $f$ be $L$-smooth and twice continuously differentiable.
Let $x^\star_u$ be an unstable saddle, and $x^\star_m$ be an almost strict local maximum. 
Then, the iterates $x_t$ of AutoGD satisfy
\[
  \P\lt(\lim_{t\to\infty} x_t = x^\star_u\rt) = 0, 
\]
and for any subsequence $t_i$, we have 
\[
  \P\lt(\lim_{i\to\infty} x_{t_i} = x^\star_m\rt) = 0.
\]
\ethm

Under some additional assumptions, we can guarantee that AutoGD converges to one 
of the local minima of $f$, instead of another critical point.
Let $\scU$ be the set of unstable saddles of $f$,
and $\scM$ the set of local minima of $f$.
We say that a set of points $\scX$ is \emph{identifiable} if $|\scX|=|f(\scX)|<\infty$.
Note that the following \cref{assum:allidentifiable} is stronger than 
\cref{assum:unstableidentifiable}.

\bassum
\label{assum:criticaltypes}
All of the critical points of $f$ are local minima, almost strict local maxima, or unstable saddles.
\eassum

\bassum
\label{assum:unstableidentifiable}
$\scU$ is identifiable, and furthermore $f(\scM)\cap f(\scU) = \emptyset$.
\eassum

\bassum\label{assum:allidentifiable}
$\scU \cup \scM$ is identifiable.
\eassum

\bthm
\label{thm:tolocalmin}
Let $f$ be twice continuously differentiable and $L$-smooth. 
Suppose further that $f$ has compact sublevel sets and satisfies \cref{assum:criticaltypes}.
Let $\scM$ be the set of local minima of $f$. The iterates $x_t$ of AutoGD satisfy:
\bitem
\item If $x_t\to x^\star$, then $x^\star\in\scM$ almost surely.
\item If $f$ satisfies \cref{assum:unstableidentifiable}, then $\min_{z\in \scM}\|x_t - z\| \to 0$ almost surely.
\item If $f$ satisfies \cref{assum:allidentifiable}, then $x_t\to x^\star \in \scM$ almost surely.
\eitem
\ethm

Finally, we obtain the asymptotic rate of convergence of AutoGD to a local minimum 
under the assumption of local (not global) strong convexity and Lipschitz 
smoothness around the optimum.
We define $\lambda_\text{min}$ and $\lambda_\text{max}$ to be the minimum and maximum 
real eigenvalues of a Hermitian matrix.

\bthm
\label{thm:localminrate}
Suppose $x_t\to x^\star$ and that there exists an $\eps_0 > 0$ such that 
such that $\|x - x^\star\| \leq \eps_0$ implies 
$f$ is twice differentiable at $x$ and 
$\lambda_\text{min}(\nabla^2 f(x)) \geq \mu^\star$ and   
$\lambda_\text{max}(\nabla^2 f(x)) \leq L^\star$. 
Then, for any $0 < \eps < \mu^\star$, the AutoGD iterates satisfy
\[
  f(x_t) = O\lt(\lt( 1 - \frac{4(c-1)(c^2-c+2)}{(c^2+1)^2} \cdot 
    \frac{\mu^\star - \eps}{L^\star + \eps}\rt)^{t/2}\rt).
\]
\ethm


\subsection{Nonasymptotics} 
\label{sec:nonasymptotics}

We conclude with a nonasymptotic result for the convergence behaviour of 
AutoGD. 
We assume unimodality of $f$ in a given gradient direction.

\bassum 
\label{assum:unimodalgamma}
For all $x\in\reals^d$, the function $g(\gamma) = f(x-\gamma \grad f(x))$, $\gamma\geq0$ 
is unimodal: there exists a $\gamma^\star(x) \in [0, +\infty]$ such that $g(\gamma)$ is 
nonincreasing for $0 \leq \gamma \leq \gamma^\star(x)$, and
nondecreasing for all $\gamma > \gamma^\star(x)$.
\eassum

We remark that \cref{assum:unimodalgamma} is satisfied in the case where $f$ is 
differentiable and convex, strongly convex, or quasi-convex, as well as for other function classes. 
Our nonasymptotic convergence result under this assumption is stated below,
which establishes that AutoGD is able to recover the theoretically optimal 
convergence rate (up to a constant) \cite[Example 1.3]{deklerk2017worst} without 
any knowledge of the smoothness constant $L$ or the strong convexity or P\L{} constant $\mu$.

\bthm
\label{thm:nonasymptotic}
Suppose $f$ is $L$-smooth, satisfies \cref{assum:unimodalgamma}, 
and $t > t_0 = \lt\lceil\lt|\log_c(\gamma_0/\slgamma)\rt|\rt\rceil$,
where 
\[
  \slgamma = \frac{2(c-1)}{L(c^2+1)} > 0.
\]
Then AutoGD satisfies
\[
\min_{\tau=0,\dots,t} \|\grad f(x_{\tau})\|^2 &\leq
\lt(\frac{L (c^2+1)^2}{(c-1)(c^2-c+2)}\rt) \cdot \frac{f(x_0)}{t-t_0}.
\]
If $f$ is also $\mu$-P\L{}, then
\[
f(x_t) &\leq f(x_{0})\lt(1 - \frac{4(c-1)(c^2-c+2)}{(c^2+1)^2}\frac{\mu}{L}\rt)^{\frac{t-t_0}{2}}.
\]
\ethm

We note that the theoretical rate is optimized when $c = 1 + \sqrt{2} \approx 2.41$, 
although with only a small improvement relative to using $c=2$. 
Due to this negligible difference and considering that \cref{thm:nonasymptotic} provides an upper 
bound on convergence, we recommend using $c=2$ in practice.

\section{EXPERIMENTS} 
\label{sec:experiments}

We assess the performance of AutoGD on a wide array of problems 
(see \cref{sec:supp_experiments} for details), including: 
26 classical optimization objectives \cite{surjanovic2013virtual,more1981testing},
61 variational inference optimization problems \cite{giordano2024dadvi}, 
and three extreme problems to test the robustness of each of the methods. 
We consider four main optimizers: 
AutoGD with a grid of initial learning rates, 
GD with a grid of constant learning rates,
backtracking line search with a grid of maximum learning rates,
and AdGD2 (Algorithm 2 in \cite{malitsky2024adaptive}) using 
a one-time line search to find an initial learning rate as suggested by the authors. 
We omit AdGD \cite{malitsky2019adgd} as preliminary results suggested similar performance 
to AdGD2 and the latter typically allows for larger learning rates.

We also consider extensions of AutoGD applied to 
both BFGS and L-BFGS (AutoBFGS and AutoLBFGS), although establishing theoretical guarantees 
for these methods is left for future work. 
These algorithms use an AutoGD step in the (L)BFGS direction. 
If we take the ``no movement'' option, we do not update the inverse Hessian approximation and we
reduce the learning rate as usual. Otherwise, we perform the usual (L)BFGS update to the inverse Hessian. 
The pseudocode for Auto(L)BFGS is given in \cref{alg:autobfgs,alg:autolbfgs}.

In our experiments we are primarily interested in 
the number of iterations that it takes for each method to reach an error tolerance and how
this number of iterations depends on the initial learning rate.
The classical optimization experiments are performed in Julia
\cite{bezanson2017julia} and remaining varational inference experiments are 
done in Python using the \texttt{dadvi} package \cite{giordano2024dadvi}. 
Experiments are performed on the ARC Sockeye compute cluster at the University 
of British Columbia.

\subsection{Classical optimization problems}

The classical optimization problems come from a standard test set for unconstrained optimization 
\cite{surjanovic2013virtual,more1981testing}, which covers optimization landscapes 
with difficult geometry, multiple local minima, and other scenarios where 
traditional optimizers could struggle.
The dimensions of the targets range from $d=2$ to $d=100$.

\begin{figure*}[!t]
  \centering
  \begin{subfigure}{0.48\textwidth}
    \centering
	\includegraphics[width=\textwidth]{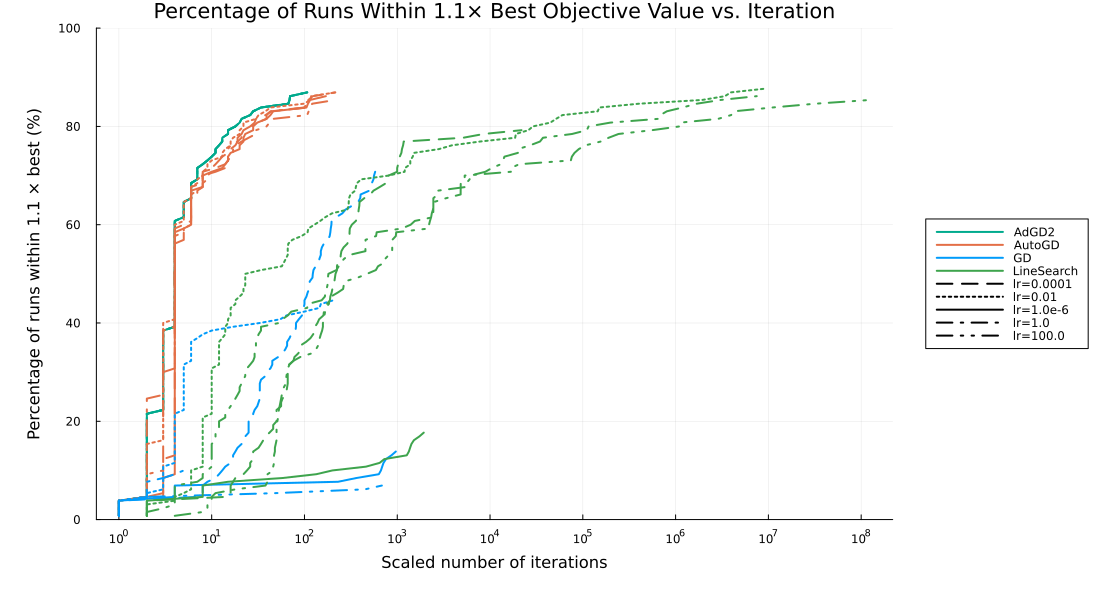}
  \end{subfigure}
  \begin{subfigure}{0.48\textwidth}
    \centering
	\includegraphics[width=\textwidth]{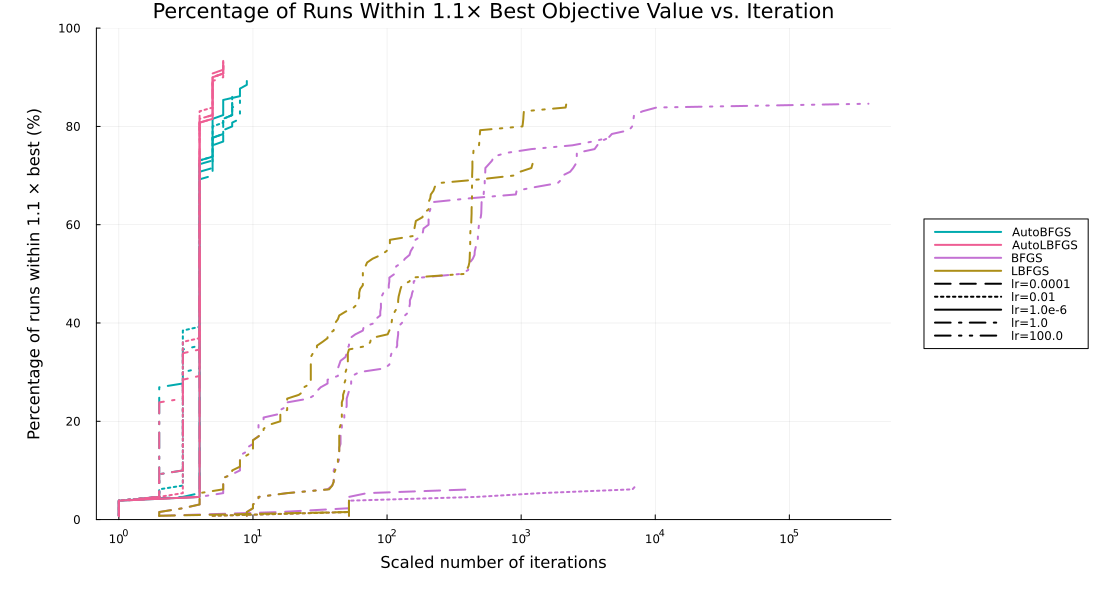}
  \end{subfigure}
	\caption{
        Percentage of runs for a given (learning rate, optimizer) combination 
        that reach within a 1.1x level of tolerance to the best 
        objective function value on the classical optimization test set. 
        \textbf{Left:} First-order methods.
        \textbf{Right:} Second-order methods.
    }
	\label{fig:classical_success_curves}
\end{figure*}

We run each combination of seed, learning rate, optimizer, and model for 
100,000 iterations. For a given learning rate and optimizer combination, at 
each time point we record the fraction of runs (over seeds and models) that 
reach a pre-specified error tolerance.
We assess whether a given optimizer achieves an objective value within a $1.1$
factor to the best objective value, where
the best value is calculated by taking the minimum obtained value across 
all seeds, optimizers, and learning rates for a given model.
The iteration time for line search-based methods is scaled to be proportional to the 
number of function/gradient evaluations as this is typically the main computational bottleneck.

The results for the first-order methods (GD, line search, AutoGD, and AdGD2) 
and second-order methods (BFGS, L-BFGS, AutoBFGS, and AutoLBFGS) are presented 
in \cref{fig:classical_success_curves}. 
Separate figures for each model and learning rate are presented in 
the supplement.

From the results among the first-order methods, we can see that AutoGD and AdGD2 
greatly outperform backtracking line search and standard gradient descent. 
Line search methods may require many function evaluations to converge 
when the maximum learning rate is set to be too large or too small, and so these 
methods are not robust in this regard. 
This is because the backtracking or bisection procedure requires a number of 
function evaluations proportional to the logarithm of the size of the search window. 
For large windows, this can amount to 10 or more function evaluations at each iteration.
Further, standard gradient descent is not able to converge 
within the given computational budget for a large fraction of the problems,
making it unreliable from a user perspective: either we have to get the learning 
rate just right, or run with an incorrect learning rate for a very long amount of time.
In contrast, we see that AutoGD is robust to the choice of initial learning rate 
and that it has performance similar to AdGD2 on these tasks. 

Among the second-order methods, we find that allowing BFGS and L-BFGS to tune 
the learning rate using a search akin to AutoGD allows for a substantial performance improvement. 
This is clearly visible in the right panel of \cref{fig:classical_success_curves}, 
where Auto(L)BFGS converges in only a few iterations for a large fraction 
of the runs.

\subsection{Variational inference problems}

We next consider a collection of black-box variational inference problems.
The goal is to fit a multivariate normal approximation to a Bayesian posterior 
using a sample average approximation (SAA) of the reverse KL divergence. 
The number of samples is fixed to $m=30$ and we perform deterministic automatic 
differentiation variational inference (DADVI)
(see \cite{giordano2024dadvi} for a justification). 
Second-order methods are omitted for this set of experiments as first-order 
optimizers are able to converge within a few (e.g., 100 to 1000) iterations.

\begin{figure}[!t]
  \centering
	\includegraphics[width=0.48\textwidth]{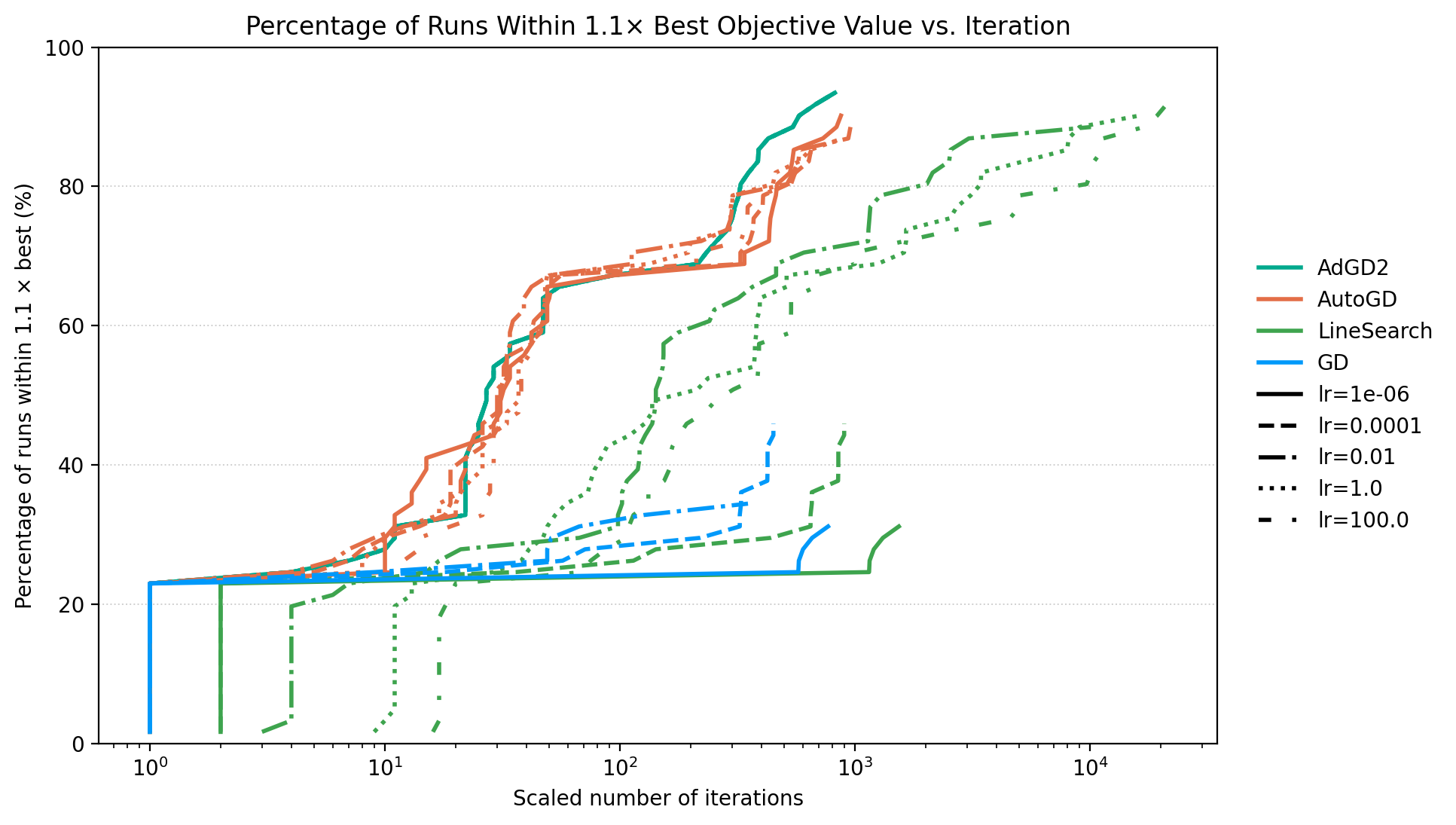}
	\caption{
        Percentage of runs for a given (learning rate, optimizer) combination 
        that reach within a 1.1x level of tolerance to the best 
        objective function value (higher is better) for various variational inference problems. 
    }
	\label{fig:dadvi_success_curves}
\end{figure}

The results for these simulations are presented in \cref{fig:dadvi_success_curves}. 
The takeaways are similar to those for the classical optimization problems: 
line search methods can require many function evaluations to converge, and 
AutoGD is robust to the choice of initial learning rate.

\subsection{Extreme objective functions}
\label{sec:extreme_examples}

\begin{figure*}[!t]
  \centering
  \begin{subfigure}{0.32\textwidth}
    \centering
	\includegraphics[width=\textwidth]{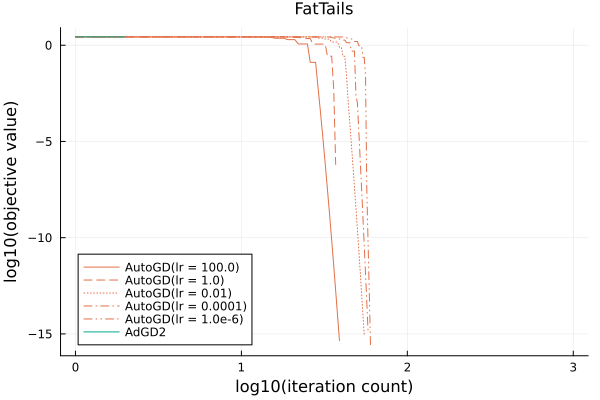}
  \end{subfigure}
  \begin{subfigure}{0.32\textwidth}
    \centering
	\includegraphics[width=\textwidth]{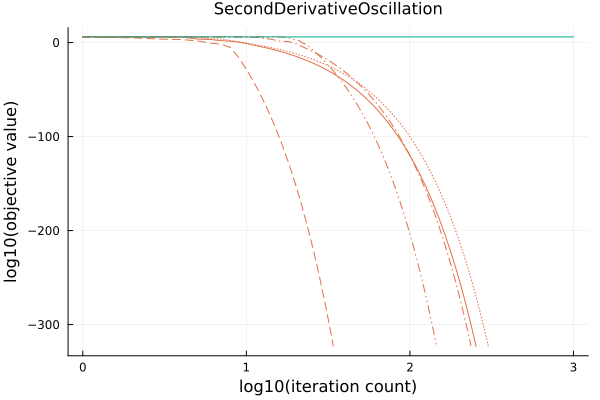}
  \end{subfigure}
  \begin{subfigure}{0.32\textwidth}
    \centering
	\includegraphics[width=\textwidth]{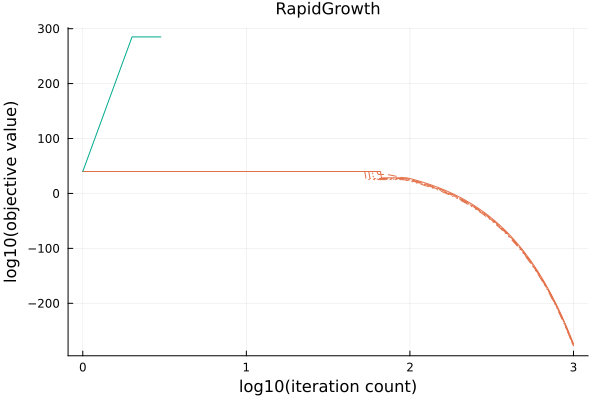}
  \end{subfigure}
  \begin{subfigure}{0.32\textwidth}
    \centering
	\includegraphics[width=\textwidth]{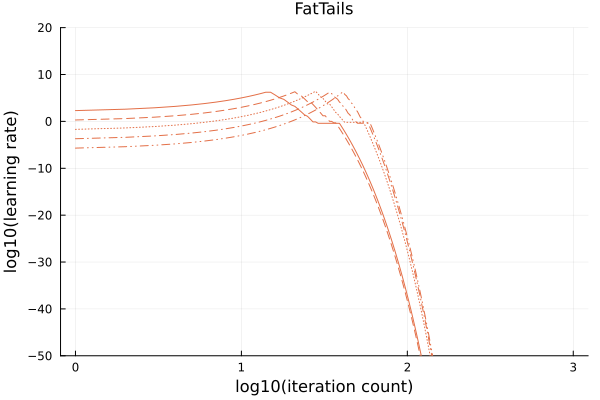}
  \end{subfigure}
  \begin{subfigure}{0.32\textwidth}
    \centering
	\includegraphics[width=\textwidth]{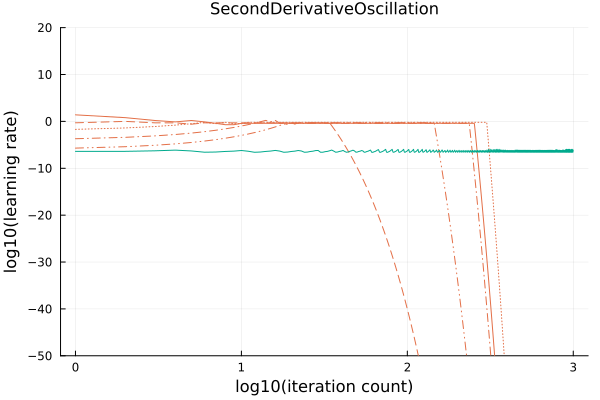}
  \end{subfigure}
  \begin{subfigure}{0.32\textwidth}
    \centering
	\includegraphics[width=\textwidth]{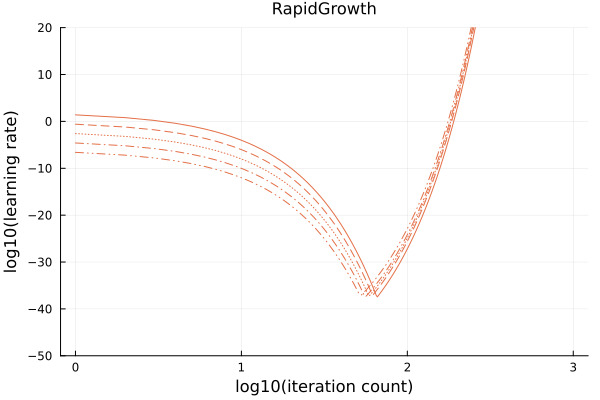}
  \end{subfigure}
	\caption{
        Performance of AutoGD and AdGD2 on various difficult objective functions.
        Objective function values (on log scale) are presented in the top row
        and the corresponding selected learning rates are in the bottom row.
        \textbf{Left:} Function with fat tails, $f(x) = \log(\log(1+x^2)+1)$.
        Optimizers are initialized at $x = 1000$.
        \textbf{Middle:} Function with rapidly changing second derivatives, 
        $f(x) = x^2 + 0.9(1-\cos(x^2))$.
        Optimizers are initialized at $x = 1000$.
        \textbf{Right:} Function with rapid growth in the tails, $f(x)=x^{20}$.
        Optimizers are initialized at $x = 100$.
    }
	\label{fig:adgd_pathologies}
\end{figure*}

Since it is desirable to have an optimizer that performs well for an extremely 
broad class of problems, we include the following examples as edge cases to test 
the robustness of all methods.
We consider the following three functions, which act as representatives from 
general classes of difficult functions that can be encountered in practice. 

\textbf{Fat tails.} For quasi-convex objectives, it is possible to have fat tails, 
such as in 
\[
  f(x) = \log(\log(1+x^2)+1). 
\]
When initialized in the tails of such functions, methods such as AdGD and AdGD2 
may struggle to learn an appropriate learning rate or encounter numerical instabilities 
due to estimates of the curvature being based on finite differencing techniques.

\textbf{Rapidly changing second derivatives.}
Because methods such as AdGD and AdGD2 rely on estimates of the local curvature to inform 
the choice of learning rate, they may struggle on objective functions where the second derivative 
changes rapidly. As a model of this behaviour, we consider 
\[
  f(x) = x^2 + 0.9(1-\cos(x^2)). 
\]

\textbf{Rapid growth.} Estimation of an appropriate learning rate may prove difficult 
for AdGD/AdGD2 when the objective function exhibits rapid growth in the tails. 
As a simple representative from this class of functions, we consider the 
polynomial $f(x) = x^{20}$.
This function also has the property of rapid decay in the interval $(-1, 1)$, 
suggesting that the learning rate should vary by several orders of magnitude upon 
entering this basin.

In \cref{fig:adgd_pathologies} we present the performance of AutoGD with various 
initial learning rates and AdGD2. For all three objective functions, 
AdGD2 either diverges, remains stagnant, or runs into numerical issues. 
In contrast, AutoGD is able to reach the minimum for all objectives and with 
all choices of initial learning rates.
In the left panel for the function with fat tails, we see that AutoGD increases 
the learning rate by several orders of magnitude and then decays the learning rate 
once it reaches the basin. 
In the middle panel with rapid oscillations of the second derivative, we see that 
AutoGD keeps an almost fixed learning rate; the general shape of $f(x) = x^2 + 0.9(1-\cos(x^2))$ 
is dominated by $x^2$ and so AutoGD learns to ignore the noise in the second derivative. 
Finally, in the right panel we see that AutoGD successfully decays the learning rate 
when initialized in the tails of a function with rapid growth. It then starts 
to increase the learning rate again once it reaches the interval $(-1,1)$ of $f(x)=x^{20}$, 
where the function exhibits rapid decay.
\section{DISCUSSION} 
\label{sec:discussion}

In this paper we presented AutoGD, an algorithm for gradient descent 
with automatic selection of learning rates. 
In our experiments, we found that AutoGD is robust to the choice of initial 
learning rate and outperforms or is on par with its competitors
on many optimization problems.
In particular, AutoGD is extremely robust to a wide variety of challenging 
objective functions including problems with heavy tails, rapid growth, and 
rapid oscillations in the local curvature.
The experimental results are also supported by convergence theory under weak assumptions,
which establish that AutoGD converges at the optimal rate (up to a constant) for $L$-smooth 
and $\mu$-strongly convex functions even when the smoothness constant is unknown.
For future work, extensions to a learning rate proposal grid with more than three elements 
would raise interesting questions about an optimal number of proposals and the choice 
of spacing between learning rates. 
Such grids would also increase the potential for parallel computing of the different learning 
rates.
Finally, while we observed empirical success of the Auto(L)BFGS methods, 
establishing convergence properties of these second-order methods 
is another potentially fruitful direction for future work.

\subsubsection*{Acknowledgements}
ABC and TC acknowledge the support of an NSERC Discovery Grant 
and a CANSSI CRT Grant.
NS acknowledges the support of a Four Year Doctoral Fellowship from the 
University of British Columbia. 
We additionally acknowledge use of the ARC Sockeye computing platform from the 
University of British Columbia.

\bibliographystyle{plain}
\bibliography{main.bib}

\clearpage
\appendix
\thispagestyle{empty}

\onecolumn
\aistatstitle{
  AutoGD: Automatic Learning Rate Selection for Gradient Descent \\
  Supplementary Materials
}

\section{Definitions} 
\label{sec:supp_definitions} 

A collection of definitions are listed below for completeness.

\bdefn{\emph{(Convex function)}} 
\label{def:convex}
A function $f : \reals^d \to \reals$ is \emph{convex} if for any $x,y \in \reals^d$ and $t \in [0,1]$ 
we have 
\[
  f(tx + (1-t)y) \leq t f(x) + (1-t) f(y).
\]
\edefn

\bdefn{\emph{(Strongly convex function)}} 
\label{def:strongly_convex}
A function $f : \reals^d \to \reals$ is \emph{$\mu$-strongly convex} for some $\mu > 0$ 
if for any $x,y \in \reals^d$ and $t \in (0,1)$ we have 
\[
  f(tx + (1-t)y) + \mu \frac{t(1-t)}{2} \|x - y\|^2 \leq t f(x) + (1-t) f(y).
\]
\edefn

\bdefn{\emph{(Polyak--\L{}ojasiewicz function)}} 
\label{def:PL}
A differentiable function $f : \reals^d \to \reals$ is \emph{$\mu$-Polyak--\L{}ojasiewicz (P\L{})} 
for some $\mu > 0$ if it is bounded from below and for any $x\in \reals^d$ we have 
\[
  f(x) - \inf f \leq \frac{1}{2\mu} \|\nabla f(x)\|^2.
\]
\edefn

\bdefn{\emph{(Smooth function)}} 
\label{def:smooth}
A differentiable function $f : \reals^d \to \reals$ is \emph{$L$-smooth} for some $L \geq 0$ 
if for any $x,y \in \reals^d$ we have 
\[
  \|\nabla f(x) - \nabla f(y)\| \leq L \|x - y\|.
\]
\edefn

\bdefn{\emph{(Local minimum)}}
\label{defn:localmin}
A \emph{local minimum} is a critical point $x^\star\in\reals^d$ such that
there exists a neighbourhood $x^\star \in U$ with 
\[
  \forall x \in U,\quad f(x) \geq f(x^\star).
\]
\edefn

\bdefn{\emph{(Identifiable points)}}
\label{defn:unique}
A set of points $\scX$ is \emph{identifiable} (with respect to $f$) if $|\scX|=|f(\scX)|<\infty$.
\edefn

\clearpage
\section{Proofs} 
\label{sec:autogd_proofs}


\subsection{AutoGD counterexamples}

\bprfof{\cref{result:autogd_diverges}}
We show $\abs{x_{t+1}} > c \abs{x_t}$ and $\gamma_t = c^{-t} \gamma_0$ 
for all $t \geq 0$. We proceed by induction. By assumption, we have that 
\[
  \abs{x_0} > \max\left\{1, \frac{c}{2p \gamma_0}, \frac{c(c+1)}{2p \gamma_0}\right\}.
\]
Then note that
\[
  x_1 
  = x_0 - \gamma \nabla f(x_0) 
  = x_0 \left(1 - 2p \gamma x_0^{2(p-1)}\right)
\]
for some $\gamma \in \{c^{-1}\gamma_0, \gamma_0, c\gamma_0\}$.
Therefore, for any $\gamma \in \{c^{-1}\gamma_0, \gamma_0, c\gamma_0\}$,
\[
  \abs{x_1} 
  = \abs{x_0} \cdot \left(2p \gamma x_0^{2(p-1)} - 1\right) 
  \geq \abs{x_0} \cdot \left(2p \gamma \abs{x_0} - 1\right) 
  > \abs{x_0} \left( c+1 - 1\right) 
  = c \abs{x_0}. 
\]
Therefore, we also have $\gamma_1 = c^{-1} \gamma_0$. This concludes the base case. 

Consider now any $t \geq 1$ and suppose that $\abs{x_i} > c \abs{x_{i-1}}$ and that $\gamma_i = c^{-i} \gamma_0$
for all $0 \leq i \leq t$. 
This implies that $\abs{x_i} > c^{i} \abs{x_0}$ for any $i \leq t$.
Therefore, for any $\gamma \in \{c^{-1}\gamma_t, \gamma_t, c\gamma_t\}$,
\[
  2p \gamma x_t^{2(p-1)} 
  &\geq 2 p c^{-(t+1)} \gamma_0 x_t^{2(p-1)} \\
  &\geq 2 p c^{-(t+1)} \gamma_0 \abs{x_0}^{2(p-1)} c^{2t(p-1)} \\
  &= 2 p c^{-1} \gamma_0 \abs{x_0} \cdot c^{-t} \abs{x_0}^{2p-3} c^{2t(p-1)} \\
  &> (c+1) \cdot c^{-t} c^{2t(p-1)} \\
  &> c+1, 
\]
and so 
\[
  \abs{x_{t+1}}
  = \abs{x_t} \cdot \left( 2p\gamma x_t^{2(p-1)} - 1 \right) 
  > c \abs{x_t},
\]
and $\gamma_{t+1} = c^{-1} \gamma_t = c^{-(t+1)} \gamma_0$.
This completes the proof.
\eprfof


\bprfof{\cref{prop:counterexample_1}}
Consider the function
\[
f(x) = |x|^{1+\eps} + b\exp\lt(-x^2\rt), \qquad b >0, \quad 0<\eps<1.
\]
Note that $f$ takes the shape of a bowl centered at $x=0$ with a bump in the middle.
To guarantee that AutoGD does not find a stationary point, we will first find a setting of
$x_0,\gamma_0,b,\epsilon$ such that AutoGD \emph{without} the ``no movement'' option satisfies $(x_t,\gamma_t) = (\pm \stx, \stgamma)$
for some $\stx,\stgamma > 0$,
tick-tocking back and forth between $\pm\stx$ forever. Then we will show that the fixed cycle
is stable, in the sense that if we instead initialize at $x_0 = \stx+\delta$ for some sufficiently small $\delta>0$,
AutoGD \emph{with} the ``no movement'' option will slowly converge to $\pm \stx$, with small decrements in each
iteration to avoid triggering the ``no movement'' option.

To begin, note that
\[
\grad f(x) &= (1+\eps)|x|^{\eps}\sgn(x) -2xb\exp\lt(-x^2\rt).
\]
In order to find a cycle, we need to find a pair $\stx, \stgamma$ satisfying
\[
-\stx = \stx - \stgamma \grad f(\stx).
\]
Note that $\forall x$,  $\grad f(-x) = -\grad f(x)$, and so finding such an $\stx$ will also guarantee that 
$\stx = -\stx - \stgamma\grad f(-\stx)$, completing the two-iteration cycle.
Substituting $\grad f(\stx)$ and solving for $\stgamma$ yields 
\[
 \stgamma  &= \frac{\stx^{1-\eps}}{\frac{1}{2}(1+\eps)-\stx^{1-\eps}b\exp\lt(-\stx^2\rt)}.
\]
We require the solution to be positive, which places a constraint on $b,\eps$:
\[
  0 &<\frac{1}{2}(1+\eps)-\stx^{1-\eps}b\exp\lt(-\stx^2\rt) \iff
  b < \frac{\frac{1}{2}(1+\eps)}{\stx^{1-\eps}\exp\lt(-\stx^2\rt)}.
\]
Next, we require that AutoGD (with the ``no movement'' option) will keep the
same learning rate $\gamma_{t} = \stgamma$.  There is no risk of setting
$\gamma_1 = c\stgamma$, since using $c\stgamma$ can only result in increasing
the function value.  But it is possible to pick $\gamma_1 = c^{-1} \stgamma$
when $f(0) < f(\stx)$  (when $\stx - c^{-1}\stgamma \grad f(\stx) = 0$ by the
above choice of $\stgamma$).  To prevent this situation, we will ensure that the
``bump'' in $f(x)$ around $x=0$ is large enough. In particular, we require that
\[
  f(0) &> f(\stx) \iff
  b > \stx^{1+\eps} + b\exp\lt(-\stx^2\rt) \iff
  b > \frac{\stx^{1+\eps}}{1-\exp\lt(-\stx^2\rt)}.
\]
Therefore, set 
\[
  b &= \frac{(1+\eps)\stx^{1+\eps}}{1-\exp\lt(-\stx^2\rt)}.
\]
This automatically satisfies the second constraint; the first is satisfied as
long as $\stx$ is large enough that 
\[
  b = \frac{(1+\eps)\stx^{1+\eps}}{1-\exp\lt(-\stx^2\rt)} &<\frac{\frac{1}{2}(1+\eps)}{\stx^{1-\eps}\exp\lt(-\stx^2\rt)} \iff
  \stx^{2}\exp\lt(-\stx^2\rt) <\frac{1}{2}\lt(1-\exp\lt(-\stx^2\rt)\rt).
\]
This inequality is satisfied for all $\stx \geq 5/4$.

Finally, we need to ensure that the oscillation is stable at $\stx$, in the sense that initializing 
$x_0 = \stx+\delta$, $\gamma_0 = \stgamma$ for some sufficiently small $\delta > 0$ will result in 
the even iterations converging $x_{2t} \to \stx$, and the odd iterations $x_{2t+1}\to -\stx$.
Consider the function
\[
h(x) = x - \stgamma \grad f(x).
\]
By the choice of $\stx,\stgamma$ above, we know that $h(\stx) = -\stx$.
Additionally, by the Taylor expansion of $h$, we have that for sufficiently small $\delta>0$, 
\[
  h(\stx + \delta) 
  &= h(\stx) + \der{h}{y}(x)\delta + O(\delta^2) \\
  &= -\stx + \der{h}{x}(x)\delta + O(\delta^2).
\]
Therefore, as long as $-1 < \der{h}{x}(\stx) < 0$, for sufficiently small $\delta>0$,
the iteration from $\stx+\delta$ will result in an updated state $-\stx-\eta\delta$ for $0<\eta < 1$, guaranteeing stability.
Repeating this logic to analyze the iteration starting from $-\stx$ results in the exact same condition on $\der{h}{x}$.
For stability, we therefore require that
\[
&\qquad \, \,\, -1 < 1 - \stgamma \grad^2 f(\stx) < 0\\
&\iff -1 < 1 - \stgamma \lt(\eps(1+\eps)|\stx|^{\eps-1} - 2b\exp(-\stx^2)+ 4\stx^2b\exp(-\stx^2) \rt) < 0\\
&\iff 2 >  \stgamma \lt(\eps(1+\eps)\stx^{\eps-1} - 2b\exp(-\stx^2)+ 4\stx^2b\exp(-\stx^2) \rt) > 1\\
&\iff 2 >  \lt(\frac{\stx^{1-\eps}}{\frac{1}{2}(1+\eps)-\stx^{1-\eps}b\exp\lt(-\stx^2\rt)}\rt) \lt(\eps(1+\eps)\stx^{\eps-1} - 2b\exp(-\stx^2)+ 4\stx^2b\exp(-\stx^2) \rt) > 1\\
&\iff 2 >  \frac{\eps(1+\eps) - 2\stx^{1-\eps}b\exp(-\stx^2)+ 4\stx^{3-\eps}b\exp(-\stx^2) }{\frac{1}{2}(1+\eps)-\stx^{1-\eps}b\exp\lt(-\stx^2\rt)}  > 1.
\]
As $\stx \to\infty$, $b =\Theta(\stx^{1+\eps})$. Therefore all the terms with $b$ can be made arbitrarily small, leaving the inequality
\[
2 > \frac{\eps(1+\eps)}{\frac{1}{2}(1+\eps)} > 1,
\]
which is satisfied for $\eps = 3/4$. Substituting $\eps = 3/4$ into the above formulae yields the desired result.
\eprfof


\bprfof{\cref{prop:counterexample_2}}
If $\gamma_0 = 2x_0/\grad f(x_0)$,
then 
\[
x_0 - c^{-1}\gamma_0 \grad f(x_0) &= 0\\
x_0 - \gamma_0 \grad f(x_0) &= -x_0\\
x_0 - c\gamma_0 \grad f(x_0) &= -2x_0.
\]
Note that $x_0$ is selected large enough such that for all $x \geq x_0$, $\grad f(x) > 0$,
because 
\[
  \nabla f(x) > 0 \iff     
  2x_0 (1 - b \exp(-x_0^2)) > 0.
\]
Because the function $f$ is symmetric we have 
\[
  f(x_0 - \gamma_0 \nabla f(x_0)) &= f(-x_0) = f(x_0) \\
  f(x_0 - c\gamma_0 \nabla f(x_0)) &= f(-2x_0) = f(2x_0) > f(x_0).
\]
AutoGD will then set the next state to be $\gamma_1 = c^{-1}\gamma_0$, $x_1 = 0$
as long as the Armijo condition is satisfied: 
\[
  &\qquad \, \, \,b = f(0) < f(x_0) - \eta \gamma_0 \|\grad f(x_0)\|^2\\
  &\iff b < x_0^2 + b\exp\lt(-x_0^2\rt) - \eta 2x_0\grad f(x_0)\\
  &\iff b < x_0^2 + b\exp\lt(-x_0^2\rt) - \eta 4x_0^2(1-b \exp\lt(-x_0^2\rt))\\
  &\iff b\lt(1-(1+4\eta x_0^2)\exp\lt(-x_0^2\rt)\rt) < x_0^2  - \eta 4x_0^2.
\]
This condition is satisfied by assumption. 
Since $\grad f(x_1) = 0$, $x_1$ is a stationary point and for all $t \geq 1$, 
AutoGD will set $x_t = x_1 = 0$.
\eprfof



\subsection{AutoGD basic behaviour}

In much of the analysis we will assume that $f$ is $L$-smooth. Let
\[
\label{eq:G}
  G(x, \gamma) &= \frac{\grad f(x)}{\|\grad f(x)\|}^T \int_0^1 2(1-t) \grad^2 f(x - t\gamma \grad f(x))\d t \frac{\grad f(x)}{\|\grad f(x)\|},
\]
where we define $G(x,\gamma) = 0$ anywhere $\|\grad f(x)\| = 0$.
Note that above, by $L$-smoothness, $f$ is guaranteed to be twice differentiable almost everywhere, 
and $\grad^2 f$ is equal to the Hessian of $f$ where it is defined.

\blem
\label{lem:basicdescent}
If $f$ is $L$-smooth, then for all $x\in\reals^d$, $\gamma>0$, and $x' = x - \gamma \grad f(x)$,
\[
f(x') &= f(x) - \gamma\lt(1-\frac{1}{2}\gamma G(x,\gamma)\rt)\|\grad f(x)\|^2, \qquad \sup |G(x,\gamma)| \leq L.
\]
\elem

\bprfof{\cref{lem:basicdescent}}
This is a direct application of the Taylor remainder theorem with explicit integral formulation, 
and the bound $|G(x,\gamma)| \leq \esssup_x \|\grad^2 f(x)\| \leq L$.
\eprfof

Smoothness is the only condition required to guarantee that the learning rates
do not decay arbitrarily. We note that the upper bound on $\eta$ is not necessary
for a result similar to this, but the following bound on $\eta$ should not impact the algorithm in practice 
(typically, $\eta$ should be very small and $c\approx 2$, so the bound will hold in almost all 
practical settings). The proof is cleaner and with tighter bounds (by the constant $c$), 
so we choose to impose it.

\blem
\label{lem:gammalowerbd}
Let $f$ be $L$-smooth, $\eta \leq (c+1)/(c^2+1)$, and define
\[
\slgamma = \frac{2(c-1)}{L(c^2+1)} > 0.
\]
Then the learning rate $\gamma_t$ of AutoGD satisfies
\[
\forall t &< \max\lt\{0, \lceil\log_c(\slgamma/\gamma_0)\rceil\rt\}, \quad \gamma_t < \slgamma\,\text{ and }\,\gamma_{t+1} > \gamma_t;\\
\forall t &\geq \max\lt\{0, \lceil\log_c(\slgamma/\gamma_0)\rceil\rt\}, \quad \gamma_t \geq \slgamma.
\]
Furthermore, the number $n_{t}$ of iterations  $\tau\in\{0,\dots, t-1\}$ in which 
$\gamma_{\tau+1} \geq \gamma_{\tau} \geq \slgamma$ satisfies
\[
n_{t} &\geq \frac{t - \lt\lceil\lt|\log_c(\gamma_0/\slgamma)\rt|\rt\rceil}{2}.
\]
\elem

\bprfof{\cref{lem:gammalowerbd}}
By \cref{lem:basicdescent}, we have the following bounds for the three proposed learning rates:
\[
f(x_t - c\gamma_t\grad f(x_t)) &\leq f(x_t)- c\gamma_t\lt(1-\frac{1}{2}c\gamma_t L\rt)\|\grad f(x_t)\|^2\\
f(x_t - \gamma_t\grad f(x_t)) &\geq f(x_t)- \gamma_t\lt(1+\frac{1}{2}\gamma_t L\rt)\|\grad f(x_t)\|^2\\
f(x_t - c^{-1}\gamma_t\grad f(x_t)) &\geq f(x_t)- c^{-1}\gamma_t\lt(1+\frac{1}{2}c^{-1}\gamma_t L\rt)\|\grad f(x_t)\|^2.
\]
We are guaranteed that the largest learning rate at iteration $t$ is the minimizer of the learning rate selection objective function if
\[
  c\gamma_t\lt(1-\frac{1}{2}c\gamma_t L\rt) &> \gamma_t\lt(1+\frac{1}{2}\gamma_t L\rt) \iff
  \gamma_t < \frac{2(c-1)}{L(c^2+1)} = \slgamma. 
\]
This choice is guaranteed to provide sufficient descent and be feasible 
(according to the Armijo condition) if
\[
  c\gamma_t\lt(1-\frac{1}{2}c\gamma_t L\rt) &\geq \eta c\gamma_t \iff
  \gamma_t \leq \frac{2 (1- \eta)}{Lc},
\]
which is satisfied if $\gamma_t < \slgamma$ since $\eta \leq (c+1)/(c^2+1)$.
Therefore, if $\gamma_t < \slgamma$ AutoGD is guaranteed to increase the learning rate $\gamma_{t+1} = c\gamma_t$.

Next, by \cref{lem:basicdescent}, we have the following bounds for the constant and small learning rates:
\[
f(x_t - \gamma_t\grad f(x_t)) &\leq f(x_t)- \gamma_t\lt(1-\frac{1}{2}\gamma_t L\rt)\|\grad f(x_t)\|^2\\
f(x_t - c^{-1}\gamma_t\grad f(x_t)) &\geq f(x_t)- c^{-1}\gamma_t\lt(1+\frac{1}{2}c^{-1}\gamma_t L\rt)\|\grad f(x_t)\|^2.
\]
We are guaranteed that keeping the learning rate constant at $\gamma_t$ produces more descent than $c^{-1}\gamma_t$ if
\[
  \gamma_t < \frac{2(1-c^{-1})}{L(1+c^{-2})} = \frac{2c(c-1)}{L(c^2+1)},
\]
and that keeping $\gamma_t$ constant is feasible (according to the Armijo condition) if
\[
  \gamma_t &\leq \frac{2(1-\eta)}{L}.
\]
Suppose $\gamma_t < c\slgamma = \frac{2c(c-1)}{L(c^2+1)}$. Then, since $\eta \leq (c+1)/(c^2+1)$,
\[
  \frac{2(1-\eta)}{L} \geq \frac{2c(c-1)}{L(c^2+1)},
\]
and hence $\gamma_t < c\slgamma$ satisfies both of the above inequalities.
Therefore, AutoGD will not decrease the learning rate if $\gamma_t < c\slgamma$.

Combining these two results, we have that 
after initialization, $\gamma_t$ increases for $\max\lt\{0, \lceil\log_c(\slgamma/\gamma_0)\rceil\rt\}$ 
iterations until $\gamma_t \geq \slgamma$, at which point $\gamma_t$ will never decrease below $\slgamma$.
Finally, since we know that AutoGD will not decrease the learning rate below $\slgamma$, the pigeonhole principle 
implies that of the $t-\max\lt\{0, \lceil\log_c(\slgamma/\gamma_0)\rceil\rt\}$ iterations after the initial period,
AutoGD can decrease the learning rate at most $\max\{0, \lceil \log_c (\gamma_0/\slgamma)\rceil\}$ times plus
half of the remaining iterations (because otherwise there must be a $\gamma_t < \slgamma$, which violates the lower bound).
Therefore,
\[
  n_t &\geq \frac{t-\max\lt\{0, \lceil\log_c(\slgamma/\gamma_0)\rceil\rt\} - \max\lt\{0, \lceil\log_c(\gamma_0/\slgamma)\rceil\rt\}}{2}\\
  &= \frac{t - \lt\lceil\lt|\log_c(\slgamma/\gamma_0)\rt|\rt\rceil}{2}.
\]
\eprfof

A key property of AutoGD is that each iteration preserves diffusivity; in other words,
if $x,\gamma$ are random with a nonatomic distribution, then the next iterates $x',\gamma'$ after
one step of AutoGD also have a nonatomic distribution.
To prove this result, we first establish a useful lemma. 

\blem
\label{lem:lebesgue_domination} 
Let $\nu$ be dominated by Lebesgue measure $\lambda$ on $\reals^d$ and
suppose $g:\reals^d \to \reals^d$ is continuously differentiable with nonzero Jacobian 
determinant except at a collection of points $B \subset \reals^d$ satisfying 
$\lambda(\bar B) = 0$, where $\bar B$ is the closure of $B$. Then, $g_\sharp \nu \ll \lambda$. 
\elem

\bprfof{\cref{lem:lebesgue_domination}} 
Consider any set $A \subset \reals^d$ such that $\lambda(A) = 0$. 
We seek to establish that $g_\sharp \nu(A) = 0$.
By assumption, $\lambda(\bar B) = 0$ and hence the set $\reals^d \setminus \bar B$ 
is open and dense in $\reals^d$.
($\reals^d \setminus \bar B$ is open because $\bar B$ is closed. It is dense because 
if it were not, there would be a neighbourhood containing only elements from $\bar B$, 
contradicting the assumption that $\lambda(\bar B) = 0$.)
For a given point $x \in \reals^d \setminus \bar B$, by the inverse function theorem there exists 
a $U_x \subset \reals^d \setminus \bar B$ such that $x \in U_x$ and $g$ is 
bijective with differentiable inverse on $U_x$. 
We can further constrain the neighbourhoods such that $\bar U_x$ is compact.
Therefore, $\{U_x : x \in \reals^d \setminus \bar B\}$ is an open covering of 
$\reals^d \setminus \bar B$.
Because $\reals^d \setminus \bar B$ is Lindel\"{o}f, there exists a countable subcovering 
of $\reals^d \setminus \bar B$, consisting of $\{U_x : x \in \scX\}$, 
where $\scX \subset \reals^d \setminus \bar B$ is countable.
Then, since $\nu \ll \lambda$, 
\[
  g_\sharp \nu(A) 
  &= \nu(g^{-1}(A)) 
  = \int_{g^{-1}(A)} \nu(x) \, \d x 
  = \int_{g^{-1}(A) \cap (\reals^d \setminus \bar B)} \nu(x) \, \d x 
  \leq \sum_{y \in \scX} \int_{g^{-1}(A) \cap U_y} \nu(x) \, \d x.
\]
Because $g$ is continuously differentiable, injective, and with nonzero Jacobian 
restricted to each $U_y$, we have
\[
  \int_{g^{-1}(A) \cap U_y} \nu(x) \, \d x 
  &= \int_{g(g^{-1}(A) \cap U_y)} \nu(g^{-1}(x)) \cdot \abs{J_{g^{-1}}(x)} \, \d x \\
  &\leq \int_{A \cap g(U_y)} \nu(g^{-1}(x)) \cdot \abs{J_{g^{-1}}(x)} \, \d x \\
  &\leq \int_{A \cap g(\bar U_y)} \nu(g^{-1}(x)) \cdot \abs{J_{g^{-1}}(x)} \, \d x \\
  &= 0,
\]
since $\abs{J_{g^{-1}}(x)}$ is bounded on $A \cap g(\bar U_y)$ (because $\bar U_y$ is compact and 
so $g(\bar U_y)$ is also compact) and $\lambda(A) = 0$. 
Therefore,
\[
  g_\sharp \nu(A) 
  &\leq \sum_{y \in \scX} \int_{g^{-1}(A) \cap U_y} \nu(x) \, \d x 
  = 0.
\]
We know $g_\sharp \nu(A) \geq 0$, and so we conclude that $g_\sharp \nu(A) = 0$.
\eprfof

\blem
\label{lem:diffuse}
Let $f$ be twice continuously differentiable and let 
$g:\reals^d\times \reals \to \reals^d\times \reals$ be a single iteration of AutoGD applied to $f$.
Suppose $x, \gamma$ have distribution $\nu$ dominated by the Lebesgue measure
on $\reals^d\times \reals$. Then the pushforward $g_\sharp\nu$ is also dominated by
the Lebesgue measure.
\elem

\bprfof{\cref{lem:diffuse}}
We partition $\reals^d\times \reals$ into 4 sets:
\[
  A_1 &= \{(x,\gamma) : g(x,\gamma) = (x-c\gamma\grad f(x),c\gamma)\}\text{ (increase)}\\
  A_2 &= \{(x,\gamma) : g(x,\gamma) = (x-\gamma\grad f(x), \gamma)\}\text{ (constant)}\\
  A_3 &= \{(x,\gamma) : g(x,\gamma) = (x-c^{-1}\gamma\grad f(x), c^{-1}\gamma)\}\text{ (decrease)}\\
  A_4 &= \{(x,\gamma) : g(x,\gamma) = (x, c^{-2}\gamma)\} \text{ (reject)}.
\]
Since these sets can be described via inequalities involving only measurable functions, they are measurable.
Consider the measures $\nu_1,\dots,\nu_4$ formed by the restriction of $\nu$ onto each of these sets,
and the maps $g_1,\dots,g_4$ formed by the restriction of $g$ on each of these sets.
We will study the sets $B_i$ on which each $g_i$ fails to be continuously differentiable 
with nonzero Jacobian determinant, and show that $\lambda(\bar B_i) = 0$.
Note that for $i\in\{1,2,3\}$,
\[
\lt|\det \grad g_i(x,\gamma)\rt| &= \det \bbmat I - s_i\gamma \grad^2 f(x) & -s_i\grad f(x)\\ 0 & s_i\ebmat 
= s_i \lt|\det \lt(I - s_i\gamma \grad^2 f(x)\rt)\rt|,
\]
where $s_1=c$, $s_2 = 1$, $s_3 = c^{-1}$, and for $i=4$, we have
$\lt|\det \grad g_4(x,\gamma)\rt| = c^{-2}$.
Therefore, $B_4 = \emptyset$ since $c^{-2} \neq 0$. 
Note that then $\lambda(\bar B_4) = 0$.
For $i=1,2,3$, denoting the eigenvalues of $\grad^2 f(x) = \{\lambda_1,\dots,\lambda_d\}$,
note that 
\[
 \{\gamma \in\reals : \det (I-s_i\gamma \grad^2 f(x)) = 0\} = 
\{\gamma \in \reals : \gamma = 1/(\lambda_j s_i) \text{ for some }j=1,\dots, d\},
\]
which is finite.
Therefore,
\[
  B_i := \cup_{j=1}^d \{(x, 1/(\lambda_j(x) s_i)) : x \in \reals^d \}.
\]
Because $f$ is twice continuously differentiable, the eigenvalues 
$\lambda_1(x), \ldots, \lambda_d(x)$ are continuous functions of $x$.
Therefore, $B_i = \bar B_i$ because the graph of a continuous function is closed. 
By the Radon--Nikodym theorem and Fubini's theorem to choose to 
integrate over $\gamma$ first 
(we use the same symbol $\nu$ to denote a measure and density with respect to the 
Lebesgue measure),
for $i=1,2,3$,
\[
  \int_{B_i} \nu(\d x, \d \gamma) 
  = \int_{B_i} \nu(x, \gamma) \, \d \gamma \, \d x
  = \int_{\reals^d} \int_0^\infty 
    \mathbbm{1}(\gamma \in \cup_{j=1}^d \{1/(\lambda_j(x) s_i)\}) \, \nu(x, \gamma) \, \d \gamma \, \d x
  = 0.
\]
The value of the integral is zero because the set of such $\gamma$ is finite, 
which has Lebesgue measure zero. 
Therefore, $\lambda(\bar B_i) = 0$ for $i=1,2,3$, as well.

To complete the proof, recall that each $\nu_i$ is dominated by the Lebesgue measure. 
Additionally, each $g_i$ is continuously differentiable and has nonzero Jacobian 
except for at $B_i$ that satisfies the conditions of 
\cref{lem:lebesgue_domination} with $\lambda(\bar B_i) = 0$.
Therefore, each pushforward $(g_i)_\sharp \nu_i$ is dominated by the Lebesgue measure 
by \cref{lem:lebesgue_domination}.  
Hence, $g_\sharp \nu = \sum_i (g_i)_\sharp \nu_i$ is also dominated by Lebesgue measure.
\eprfof



\subsection{AutoGD asymptotics}

\bprfof{\cref{prop:monotonef}}
Since $f(x_{t+1}) \leq f(x_t)$ for all $t$ by design, and $f(x_t) \geq 0$ by assumption,
the monotone convergence theorem guarantees that there exists $\underline{f}\geq 0$ such 
that  $\lim_{t\to\infty} f(x_t) = \underline{f}$.
\eprfof

\bprfof{\cref{thm:tostationarity}}
By \cref{lem:gammalowerbd}, there exists a $T \in \nats$ such that $t\geq T$ implies
\[
\gamma_t \geq \slgamma > 0.
\]
Suppose $\grad f(x_t) \not\to 0$. Because $f$ is $L$-smooth, $0 \leq \|\nabla f(x_t)\| \leq L$ 
for all $t$. Therefore, there exists a subsequence
$t_i$ such that $\|\grad f(x_{t_i})\| \to \limsup \|\grad f(x_t)\| > 0$.
Note that at step $t_i$ we are guaranteed to reject the step and shrink the learning rate if
\[
f(x_{t_i}) - f(x_{t_i+1}) - \eta c^{-1} \gamma_{t_i}\|\grad f(x_{t_i})\|^2 &< 0.
\]
For any $\epsilon > 0$, we can pick $i$ large enough such that 
$\|\grad f(x_{t_i})\| \geq (1-\epsilon) \limsup \|\grad f(x_t)\|$
and for all $k\geq 0$, $f(x_{t_i+k}) - f(x_{t_i+k+1}) < \epsilon$ by \cref{prop:monotonef}.
We therefore have that for sufficiently large $i$,
\[
  f(x_{t_i}) - f(x_{t_i+1}) - \eta c^{-1} \gamma_{t_i}\|\grad f(x_{t_i})\|^2 &\leq
  \epsilon - \eta c^{-1} \gamma_{t_i} (1-\eps)\limsup \|\grad f(x_t)\|.
\]
Therefore, if
\[
  \gamma_{t_i} > \frac{\epsilon c}{\eta (1-\eps)\limsup \|\grad f(x_t)\|},
\]
AutoGD is guaranteed to reject the step and shrink the learning rate for $k$ iterations until $\gamma_{t_i+k}$ is below this threshold. 
By picking $\epsilon$ small enough, this guarantees that the algorithm will shrink $\gamma_t$ below $\slgamma$, which is a contradiction. 
Therefore, $\grad f(x_t) \to 0$.
\eprfof

We note that with \cref{thm:tostationarity}, the iterates may still escape to infinity. 
To guarantee that this does not occur, we impose that $f$ has compact sublevel sets.
In this situation, we know that the iterates converge to the set of critical points of $f$ at a particular level $f(x) = \slf$.

\blem\label{lem:tocritical}
Suppose $f$ is $L$-smooth and has compact sublevel sets. Then there exists $\slf \geq 0$ such that the set 
\[
D = \{x : \grad f(x) = 0, \, f(x) = \slf\}
\]
is nonempty and compact, and the iterates $x_t$ of AutoGD satisfy 
$\min_{z \in D} \|x_t-z\| \to 0$ as $t\to\infty$.
\elem
\bprf
Denote $B = \{x : f(x)=\slf\}$, and $C = \{x : \grad f(x) = 0, f(x)\leq f(x_0)\}$. 
$B$ is closed, since $f$ is continuous, and
$C$ is the intersection of a closed set and a compact set, since $f$ has compact sublevel sets and $\grad f$ is Lipschitz (and hence continuous).
Therefore $D = B\cap C$ is also compact.

By \cref{prop:monotonef,thm:tostationarity}, $f(x_t)\downarrow \slf$ and $\grad f(x_t)\to 0$, and the $x_t$ remain in the set $\{f(x)\leq f(x_0)\}$.
Since this set is compact, there exists a convergent subsequence $x_{t_i}\to x^\star$ such that $f(x^\star)\leq f(x_0)$. 
But by $L$-smoothness, $f(x^\star) = \slf$ and $\grad f(x^\star) = 0$, and so
$D$ is nonempty.

Now consider any subsequence $a_{t_i}$ of the sequence $a_t = \min_{z\in D}\|x_t-z\|$.
Since $x_{t_i}$ remains in a compact set, there exists a convergent subsequence $x_{t_{i_j}}\to x^\star$.
By the previous logic, $x^\star$ satisfies $f(x^\star)=\slf$ and $\grad f(x^\star) = 0$.
Therefore setting $z = x^\star$ in the optimization yields $a_{t_{i_j}} \to 0$.
Since any subsequence of $a_t$ contains a further subsequence converging to 0, $a_t \to 0$.
\eprf

\bprfof{\cref{lem:avoidbadcritical}}
We have that $f$ is $L$-smooth and by assumption twice continuously differentiable.
By \cref{lem:diffuse}, for any set $N$ of Lebesgue measure zero,
\[
  \P(\exists t\in\nats : x_t \in N)\leq \sum_t \P(x_t \in N) = 0.
\]
Let $x^\star$ be an almost strict local maximum and $N$ the set of Lebesgue measure 
zero used in the definition of an almost strict local maximum. Then
if there exists a subsequence $x_{t_i} \to x^\star$, we have that $f(x_{t_i}) \to f(x^\star)$. But
since $x_{t_i} \notin N$ almost surely, we have that for all $i$, $f(x_{t_i}) < f(x^\star)$ almost surely, and so convergence 
contradicts \cref{prop:monotonef}. 

Next assume $x^\star$ is an unstable saddle. If $x_t\to x^\star$, then
$\|x_t-x^\star\|^2\to 0$. So there exists some $T\in\nats$ such that $t\geq T$
implies $x_t\in U$, where $U$ is the neighourhood in the definition of an unstable saddle. 
Therefore, for all sufficiently large $t$ and with $\|v\| = 1$ from the definition of an unstable saddle,
\[
\|x_{t+1}-x^\star\|^2 &= ((x_{t+1}-x^\star)^Tv)^2 + \| (I-vv^T)(x_{t+1}-x^\star)\|^2\\
&\geq ((x_{t+1}-x^\star)^Tv)^2\\
&= ((x_t - \gamma_{t+1}\grad f(x_t) - x^\star)^Tv)^2\\
&= ((x_t-x^\star)^Tv)^2 + \gamma_{t+1}^2(\grad f(x_t)^Tv)^2 + 2\gamma_{t+1}(x^\star-x_t)^Tv v^T\grad f(x_t)\\
&\geq ((x_t-x^\star)^Tv)^2,
\]
which holds almost surely. Since $\{x : (x-x^\star)^Tv = 0\}$ has measure zero, 
with probability one we have
$((x_t-x^\star)^Tv)^2 > 0$. 
Define $a_t := ((x_t-x^\star)^Tv)^2$. By the argument above we have 
$a_{t+k} \geq a_t > 0$ for all $k \geq 1$ almost surely, 
so that $\|x_{t+k} - x^\star\|^2 \geq a_{t+k-1} \geq a_t > 0$,
which contradicts convergence.
\eprfof

\bprfof{\cref{thm:tolocalmin}}
Suppose $x_t\to x^\star$. Then by \cref{lem:avoidbadcritical}, $x^\star$ is not unstable or
an almost strict local maximum, so $x^\star$ must be a local minimum.

By \cref{lem:tocritical}, the iterates $x_t$ of AutoGD satisfy 
$\min_{z \in D} \|x_t-z\| \to 0$ as $t\to\infty$. 
\cref{lem:avoidbadcritical} ensures that the iterates do not have a local maximum
as a limit point, so 
$\min_{z \in \scU\cup\scM} \|x_t-z\| \to 0$ as $t\to\infty$. 
Suppose $f$ satisfies \cref{assum:unstableidentifiable}. Then
if any point $x^\star \in \scU$ is a limit point of the sequence,
by monotonicity (\cref{prop:monotonef}), there is a $z\in\scU$ such that $x_t\to z$.
This cannot occur by \cref{lem:avoidbadcritical}, and hence
$\min_{z \in \scM} \|x_t-z\| \to 0$ as $t\to\infty$. 
Suppose instead that $f$ satisfies \cref{assum:allidentifiable}.
Then by the same argument, there is a $z \in \scU\cup\scM$ such that $x_t\to z$,
and hence $z\in \scM$.
\eprfof

To prove our final asymptotic result, we establish a lemma that shows that 
it is possible to extend a function $f$ that is locally strongly convex and $L$-smooth 
on an $\epsilon_0$-neighbourhood to a function $g$ that is globally strongly convex and $L$-smooth. 
We keep $f$ on a smaller $\delta$-neighbourhood, and then interpolate 
between $f$ and its second-order Taylor approximation outside of this 
neighbourhood to obtain $g$.

\blem[Continuation of locally smooth, strongly convex functions]
\label{lem:f_continuation}
Define $f : \reals^d \to \reals$ and $x^\star\in\reals^d$.
Suppose there exists $\epsilon_0 > 0$ such that $\|x-x^\star\| \leq \epsilon_0$ implies that $f$ is twice differentiable with $\mu I \preceq \grad^2 f(x) \preceq L I$. 
Then there exists a nonnegative function $m:\reals_+ \to\reals_+$, $\lim_{\epsilon\to 0}m(\epsilon) = 0$
such that for all $0 \leq \delta < \epsilon \leq \epsilon_0$, 
there exists a globally twice differentiable, $L_g$-smooth, and $\mu_g$-strongly convex function $g:\reals^d \to \reals$
such that $f(x) = g(x)$ for $\|x-x^\star\|\leq \delta$, where
\[
L_g &=L  + m(\epsilon)\frac{\epsilon^4}{(\epsilon^2-\delta^2)^2}, \qquad 
\mu_g =\mu  - m(\epsilon)\frac{\epsilon^4}{(\epsilon^2-\delta^2)^2}.
\]
\elem

\bprfof{\cref{lem:f_continuation}}
Set $\delta < \epsilon$.
Define $h$ to be the second order Taylor expansion of $f$ around $x^\star$:
\[
h(x) = f(x^\star) + \grad f(x^\star)^T(x-x^\star) + \frac{1}{2}(x-x^\star)^T\grad^2 f(x^\star)(x-x^\star).
\]
Note that $h$ is globally twice differentiable, $L$-smooth, and $\mu$-strongly convex.
Define $\phi : \reals_+ \to [0,1]$ to be the function
\[
\phi(t) = 
\begin{cases} 
    0, & t \leq \delta^2\\
    s\lt(\frac{t-\delta^2}{\epsilon^2-\delta^2}\rt), & \delta^2 < t < \epsilon^2\\
    1, & \epsilon^2 \leq t,
\end{cases}, \qquad \text{where} \qquad 
s\lt(t\rt) = t^3(6t^2-15t+10).
\]
Note that $\phi$ is globally twice continuously differentiable, and continuously and monotonically increases from $0$ to $1$ on the interval $t\in[\delta^2, \epsilon^2]$.
Finally, define
\[
g(x) = \lt(1-\phi\lt(\|x-x^\star\|^2\rt)\rt)f(x) + \phi\lt(\|x-x^\star\|^2\rt)h(x).
\]
Then $g(x) = f(x)$ for $\|x-x^\star\|\leq \delta$, and $g(x) = h(x)$ for $\|x-x^\star\| \geq \epsilon$.
Furthermore,
\[
  \grad g(x) &= (1-\phi(\|x-x^\star\|^2))\grad f(x) + \phi(\|x-x^\star\|^2)\grad h(x) + 2(h(x) - f(x))\phi'(\|x-x^\star\|^2)(x-x^\star)\\
  \grad^2 g(x) &= (1-\phi(\|x-x^\star\|^2))\grad^2 f(x) +\phi(\|x-x^\star\|^2)\grad^2 h(x)\\
    &{\quad} + 2\phi'(\|x-x^\star\|^2)\lt((x-x^\star)(\grad h(x)-\grad f(x))^T+(\grad h(x) - \grad f(x))(x-x^\star)^T\rt)\\
    &{\quad}+ 2(h(x) - f(x))\lt(\phi'(\|x-x^\star\|^2)I+ 2\phi''(\|x-x^\star\|^2)(x-x^\star)(x-x^\star)^T\rt).
\]
Since $\phi(t) = 1$ for $t \geq \epsilon^2$, the second derivative of $g$ is defined everywhere (since the $\grad^2 f$ term only
needs to be defined when $\phi < 1$). Since $\phi' = \phi'' = 0$ outside of the annulus $\delta < \|x-x^\star\| < \epsilon$,
$\grad^2 g(x) = \grad^2 f(x)$ for $\|x-x^\star\|\leq \delta$,
and $\grad^2 g(x) = \grad^2 h(x)$ for $\|x-x^\star\| \geq \epsilon$, 
both of which satisfy $\mu I \preceq \grad^2 \preceq L I$.
It remains to analyze $\grad^2 g(x)$ on the annulus $\delta < \|x-x^\star\| < \epsilon$.
Note that
\[
  &\forall t\in[\delta^2,\epsilon^2], \,\, 0 \leq \phi'(t) \leq \frac{15}{8}\cdot
    (\epsilon^2-\delta^2)^{-1} \\ 
  &\max_{\delta^2 \leq t \leq \epsilon^2} |\phi''(t)| =
    \frac{10}{\sqrt{3}} (\epsilon^2-\delta^2)^{-2} \leq 6(\epsilon^2-\delta^2)^{-2}.  
\]
Therefore on the annulus, by Taylor's theorem, there exists a nonnegative function 
$m_1(\epsilon) \to 0$ as $\epsilon \to 0$ such that
\[
  \grad^2 g(x) &\succeq \lt(\mu  - 6\frac{|h(x)-f(x)|}{\epsilon^2-\delta^2}\lt(\frac{5}{8} + 4\frac{\epsilon^2}{\epsilon^2-\delta^2} \rt)\rt)I\\
  &\succeq \lt(\mu  - m_1(\epsilon)\frac{\epsilon^4}{(\epsilon^2-\delta^2)^2}\rt)I.
\]
Similarly, there exists a nonnegative function $m_2(\eps) \to 0$ as $\eps \to 0$, such that
\[
  \grad^2 g(x) &\preceq \lt(L  
  + \frac{15}{2}\frac{\epsilon}{\epsilon^2-\delta^2} \|\grad h(x)-\grad f(x)\| 
  + 6\frac{|h(x)-f(x)|}{\epsilon^2-\delta^2}\lt(\frac{5}{8} + 4\frac{\epsilon^2}{\epsilon^2-\delta^2} \rt)\rt)I\\
  &\preceq \lt(L + m_2(\epsilon)\frac{\epsilon^4}{(\epsilon^2-\delta^2)^2}\rt)I.
\]
We may then set $m(\eps) = \max \{m_1(\eps), m_2(\eps)\}$ to complete the proof.
\eprfof

\bprfof{\cref{thm:localminrate}}
Fix $0 < \eps < \mu^\star$. Let $m$ be as in \cref{lem:f_continuation}. Because 
$m(\tilde \eps) \to 0$ as $\tilde \eps \to 0$, there exists an 
$0 < \tilde\eps \leq \eps_0$ such that 
\[
  \frac{16}{9} m(\tilde\eps) \leq \eps.
\]
Setting $\delta = \tilde\eps/2 < \eps_0$, we have that 
$f$ admits an extension $g$ such that $f(x) = g(x)$ and $\nabla f(x) = \nabla g(x)$ 
for all $\|x - x^\star\| \leq \delta$, 
and $g$ is globally $(L^\star+\eps)$-smooth and $(\mu^\star-\eps)$-strongly convex. 

Next, we define a compact set  
\[
  N = \{x : g(x) \leq b \}
\]
for some $b$ such that $N \subset \{x : \|x - x^\star\| \leq \delta\}$. 
This is possible because $g$ is strongly convex and has compact sublevel sets. 
Because $x_t \to x^\star$, there exists a $T \in \nats$ such that for $t \geq T$ 
we have $x_t \in N$. 
We argue now that the iterates of AutoGD starting at $(x_T, \gamma_T)$ applied to 
$f$ are exactly the same as the iterates obtained applying the method to $g$. 
We denote these iterates as $(x_t^f, \gamma_t^f)$ and $(x_t^g, \gamma_t^g)$, respectively, 
for $t \geq T$. Our claim is then that $(x_t^f, \gamma_t^f) = (x_t^g, \gamma_t^g)$ 
for all $t \geq T$. We prove this claim by induction. 

At $t = T$, we initialize $(x_T^f, \gamma_T^f) = (x_T^g, \gamma_T^g)$ and so the base case holds.
Next, suppose that for a $t \geq T$, we have $(x_t^f, \gamma_t^f) = (x_t^g, \gamma_t^g)$. 
We show that $(x_{t+1}^f, \gamma_{t+1}^f) = (x_{t+1}^g, \gamma_{t+1}^g)$. 
Because $x_t^f = x_t^g$ and $x_t^f \in N$, we have $f(x_t^f) = g(x_t^g)$ and 
$\nabla f(x_t^f) = \nabla g(x_t^g)$. Observe that
\[
  x_{t+1}^f 
  &\in \{x_t^f - \gamma \nabla f(x_t^f) : \gamma \in \{0, c^{-1} \gamma_t, \gamma_t, c \gamma_t\}\} \\
  &= \{x_t^g - \gamma \nabla g(x_t^g) : \gamma \in \{0, c^{-1} \gamma_t, \gamma_t, c \gamma_t\}\} \\
  &\ni x_{t+1}^g.
\]
By definition of $T$, when AutoGD is applied to $f$, 
all proposals $x'$ such that $x' \notin N$ are rejected. 
Similarly, by construction of $N$, for any point 
$x'$ such that $x' \notin N$, we have
\[
  g(x')
  > \max_{x \in N} g(x)
  = \max_{x \in N} f(x).
\] 
Therefore, applying AutoGD starting at $(x_t, \gamma_t)$ on the functions $f$ and $g$
must necessarily remain in the compact set $N$. 
However, $f(x) = g(x)$ and $\nabla f(x) = \nabla g(x)$ 
whenever $x \in N$, and so we must have 
$x_{t+1}^f = x_{t+1}^g$ and $\gamma_{t+1}^f = \gamma_{t+1}^g$. 
Therefore, by induction, $(x_t^f, \gamma_t^f) = (x_t^g, \gamma_t^g) = (x_t, \gamma_t)$ 
for all $t \geq T$.

We then employ \cref{thm:nonasymptotic} to obtain the result, noting that strongly convex 
functions satisfy the unimodality property of \cref{assum:unimodalgamma}.
\eprfof


\subsection{AutoGD nonasymptotics}

\blem\label{lem:halfitersdescent}
Suppose $f$ is $L$-smooth and satisfies \cref{assum:unimodalgamma}.
Then for all $t\in\nats$, there are 
\[
n_t &\geq \frac{t - \lt\lceil\lt|\log_c(\gamma_0/\slgamma)\rt|\rt\rceil}{2}
\]
iterations $x_\tau$, $\tau \in \{0, \dots, t-1\}$  of AutoGD  such that
$f(x_{\tau+1}) \leq f(x_{\tau}) - \slgamma\lt(\frac{c^2-c+2}{c^2+1}\rt)\|\grad f(x_\tau)\|^2$,
and for the remaining iterations, $f(x_{\tau+1}) \leq f(x_\tau)$.
\elem
\bprf
Consider any step $\tau$ where $\gamma_{\tau+1}\geq\gamma_\tau\geq\slgamma$.
Since $\gamma_{\tau+1} \geq \gamma_\tau$, it must be the case that 
$\gamma_\tau \leq \gamma^\star(x_\tau)$ by \cref{assum:unimodalgamma},
and so the descent using $\gamma_\tau$ is bounded by the descent with any $\gamma \leq \gamma_\tau$.
So since $\gamma_\tau\geq\slgamma$,
\[
f(x_{\tau+1}) = f(x_\tau - \gamma_{\tau+1} \grad f(x_\tau)) &\leq f(x_\tau - \slgamma \grad f(x_\tau)).
\]
By \cref{lem:basicdescent},
\[
f(x_\tau - \slgamma \grad f(x_\tau)) &\leq f(x_\tau) - \slgamma\lt(1-\frac{1}{2}\slgamma L\rt)\|\grad f(x_\tau)\|^2\\
&= f(x_\tau) - \slgamma\lt(\frac{c^2-c+2}{c^2+1}\rt)\|\grad f(x_\tau)\|^2.
\]
By \cref{lem:gammalowerbd}, the number  of iterations $n_t$ in which $\gamma_{\tau+1} \geq \gamma_\tau\geq \slgamma$ is
at least $\frac{t - \lt\lceil\lt|\log_c(\gamma_0/\slgamma)\rt|\rt\rceil}{2}$. 
By \cref{prop:monotonef}, the remaining iterations satisfy $f(x_{\tau+1})\leq f(x_\tau)$.
\eprf

\bprfof{\cref{thm:nonasymptotic}}
Let $t_i$ be the subsequence of iterations such that $\gamma_{t_i+1}\geq \gamma_{t_i} \geq \slgamma$.
By \cref{prop:monotonef,lem:halfitersdescent},
\[
f(x_{t_{i+1}}) \leq f(x_{t_i+1}) &\leq f(x_{t_i})- \slgamma\lt(\frac{c^2-c+2}{c^2+1}\rt)\|\grad f(x_{t_i})\|^2.\label{eq:subsequenceprotobd}
\]
Denote $n_t$ to be the number of such iterations prior to iteration $t$.
Telescoping this inequality and using monotonicity of $f$ by \cref{prop:monotonef} yields
\[
  \sum_{i=1}^{n_t}\slgamma\lt(\frac{c^2-c+2}{c^2+1}\rt) \|\grad f(x_{t_i})\|^2 
  &\leq \sum_{i=1}^{n_t} f(x_{t_i}) - f(x_{t_{i+1}}) \\
  &= f(x_{t_1}) - f(x_{t_{n_t+1}}) \\ 
  &\leq f(x_{t_1}) \\
  &\leq f(x_0).
\]
Therefore,
\[
\min_{\tau=0,\dots,t} \|\grad f(x_{\tau})\|^2 
&\leq \min_{i=1,\dots,n_t}\|\grad f(x_{t_i})\|^2\\
&\leq \sum_{i=1}^{n_t} \frac{\slgamma\lt(\frac{c^2-c+2}{c^2+1}\rt)}{n_t\slgamma\lt(\frac{c^2-c+2}{c^2+1}\rt)} \|\grad f(x_{t_i})\|^2\\
&\leq \frac{f(x_0)}{n_t\slgamma\lt(\frac{c^2-c+2}{c^2+1}\rt)}.
\]
By \cref{lem:halfitersdescent},
\[
\min_{\tau=0,\dots,t} \|\grad f(x_{\tau})\|^2 &\leq
\frac{f(x_0)}{\slgamma\lt(\frac{c^2-c+2}{c^2+1}\rt)\lt(\frac{t-\lt\lceil\lt|\log_c(\gamma_0/\slgamma)\rt|\rt\rceil}{2}\rt)}\\
&=\lt(\frac{L (c^2+1)^2}{(c-1)(c^2-c+2)}\rt)\frac{f(x_0)}{t-\lt\lceil\lt|\log_c(\gamma_0/\slgamma)\rt|\rt\rceil}.
\]
If $f$ is also $\mu$-P\L{}, then \cref{eq:subsequenceprotobd} yields 
\[
f(x_{t_{i+1}}) &\leq f(x_{t_i})\lt(1 - 2\mu\slgamma\frac{c^2-c+2}{c^2+1}\rt),
\]
so therefore
\[
f(x_t) \leq f(x_{t_{n_t}}) &\leq f(x_{0})\lt(1 - 2\mu\slgamma\frac{c^2-c+2}{c^2+1}\rt)^{n_t}\\
&\leq f(x_{0})\lt(1 - \frac{4(c-1)(c^2-c+2)}{(c^2+1)^2}\frac{\mu}{L}\rt)^{\frac{t-\lt\lceil\lt|\log_c(\gamma_0/\slgamma)\rt|\rt\rceil}{2}}.
\]
\eprfof


\clearpage
\section{Additional details and results} 
\label{sec:supp_experiments}

\subsection{General details of experiments} 
\label{sec:ML_details}

In all of our experiments, the initial learning rate grid that we use for AutoGD, GD, and line search is 
$\gamma \in \{100, 1, 10^{-2}, 10^{-4}, 10^{-6}\}$. 
The initial learning rate for AdGD2 is tuned using a one-time line search approach 
as suggested in \cite{malitsky2024adaptive}. 

In our implementation of AutoGD we do not parallelize over the three function evaluations 
as our objectives can be evaluated very quickly. However, in our experimental results, 
we study the model of computational cost under the assumption that the implementation 
involves parallelization over the three learning rates in the grid.

For our initialization of AutoGD, we set $\log \gamma_0 \sim N(0, 10^{-12})$ 
and $x_0 \sim N(x_0', 10^{-12})$, where $x_0'$ is a preliminary (possibly random) 
initialization. 
For the classical optimization experiments with five different seeds, 
four out of the five seeds are drawn from a standard normal distribution for $x_0'$, whereas 
one of the seeds is based on suggestions from \cite{more1981testing,surjanovic2013virtual} or 
by guessing a reasonable initialization for the optimizers not too close to the optimum.

\subsubsection{Classical experiments} 

All classical optimization objectives can be found in 
\cite{surjanovic2013virtual,more1981testing}, except for the ``valley'' target which is a 
synthetic example with $d=2$ defined as 
\[
  f(x) = 1 - \frac{1}{1 + x_1^2 + 4x_2^2}.
\]
The selected test functions include: 
Beale ($d = 2$), 
Biggs Exp6 ($d=6$), 
Box 3D ($d=3$), 
Brown badly scaled ($d=2$), 
Brown--Dennis ($d=4$), 
Gaussian ($d=3$), 
Gulf research ($d=3$), 
Helical valley ($d=3$), 
Matyas ($d=2$), 
Penalty 1 ($d=2$), 
Penalty 1 ($d=100$), 
Penalty 2 ($d=2$), 
Penalty 2 ($d=100$), 
Powell badly scaled ($d=2$), 
Powell singular ($d=4$), 
Powell singular ($d=100$), 
Rosenbrock ($d=2$), 
Rosenbrock ($d=100$), 
Three-hump camel ($d=2$), 
Trigonometric ($d=10$), 
Trigonometric ($d=100$), 
Variably-dimensioned ($d=2$), 
Variably-dimensioned ($d=100$), 
Valley ($d=2$), 
Watson ($d=31$), and
Wood ($d=4$).

For each of these targets, we run a given optimizer and learning rate combination 
with five different seeds. 
The objectives are shifted by one, so that the minimum for (most of) the objectives is 
$\log f^\star \approx 0$.

\subsubsection{Variational inference experiments} 
All deterministic ADVI problems can be found in \cite{giordano2024dadvi}. 
We borrowed code from the \texttt{dadvi} and \texttt{dadvi-experiments} 
repos, located at 
\url{https://github.com/martiningram/dadvi}
and \url{https://github.com/martiningram/dadvi-experiments}, respectively. 
We made some small modifications to allow for first-order custom optimizers 
and for storing the relevant objective, learning rate, and backtrack traces. 


\subsection{Details of AutoGD and additional experimental results} 
\label{sec:additional_autogd_experiments}

A preliminary version of AutoGD was introduced in \cite{surjanovic2025autosgd}. 
However, the presentation in that paper was brief and did not provide many 
convergence guarantees or simulation results. 
In this work, we modify the algorithm by introducing the diffuse initialization 
and the Armijo condition, establishing several important theoretical results. 
We also perform experiments on a wide variety of problems with the modified algorithm.

\begin{figure*}[!t]
  \centering
  \begin{subfigure}{0.32\textwidth}
    \centering
    \includegraphics[width=\textwidth]{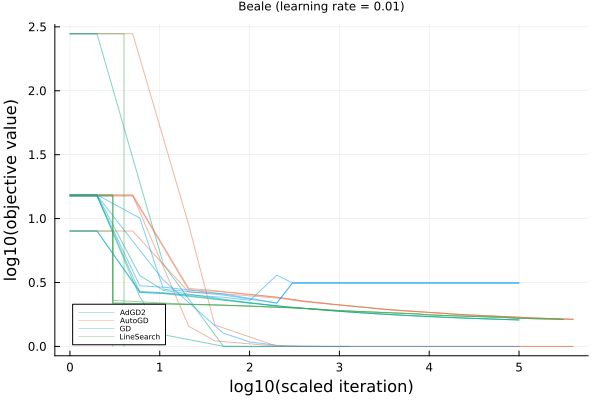}
  \end{subfigure} 
  \begin{subfigure}{0.32\textwidth}
    \centering
    \includegraphics[width=\textwidth]{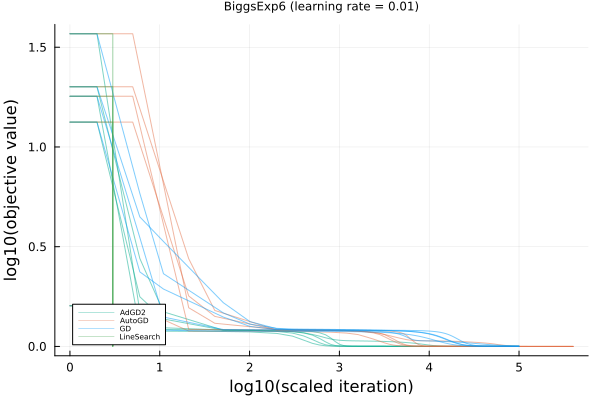}
  \end{subfigure}
  \begin{subfigure}{0.32\textwidth}
    \centering
    \includegraphics[width=\textwidth]{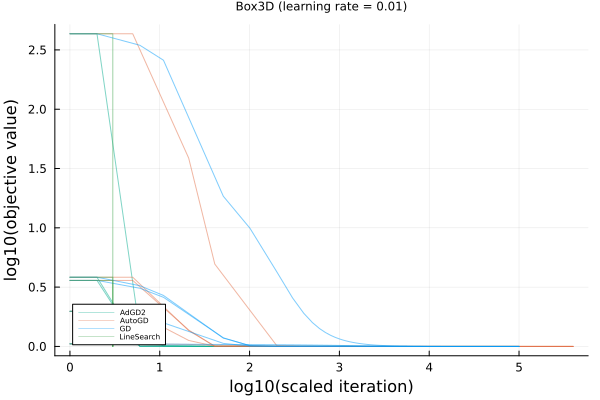}
  \end{subfigure}
  \begin{subfigure}{0.32\textwidth}
    \centering
    \includegraphics[width=\textwidth]{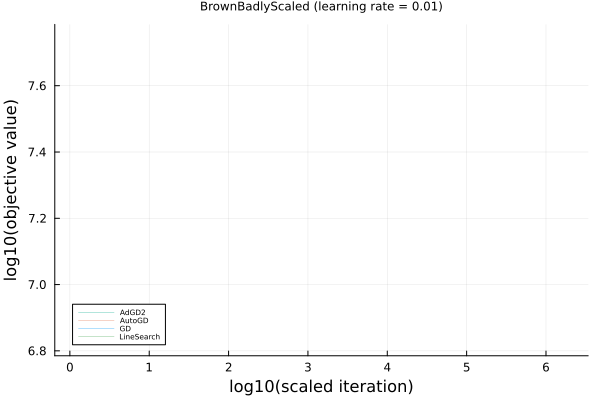}
  \end{subfigure} 
  \begin{subfigure}{0.32\textwidth}
    \centering
    \includegraphics[width=\textwidth]{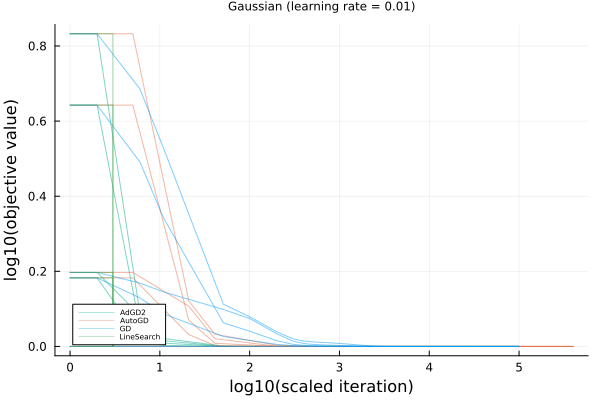}
  \end{subfigure}
  \begin{subfigure}{0.32\textwidth}
    \centering
    \includegraphics[width=\textwidth]{img/deterministic_new/deterministic_target_Gaussian_learning_rate_0.01.png}
  \end{subfigure} 
  \begin{subfigure}{0.32\textwidth}
    \centering
    \includegraphics[width=\textwidth]{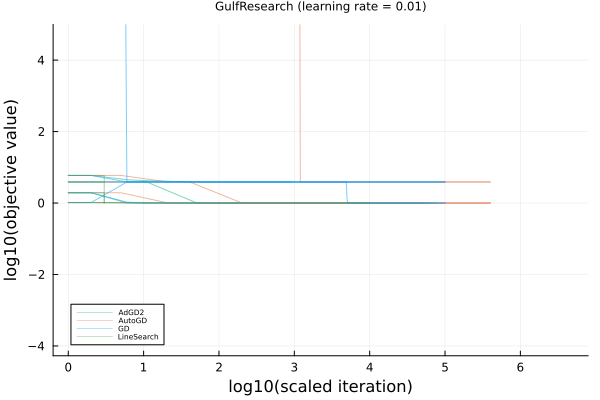}
  \end{subfigure}
  \begin{subfigure}{0.32\textwidth}
    \centering
    \includegraphics[width=\textwidth]{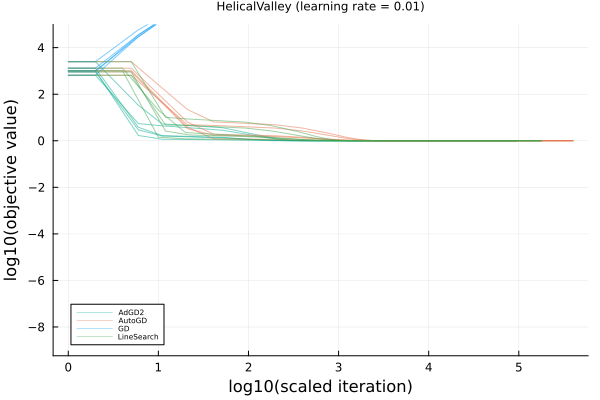}
  \end{subfigure}
  \begin{subfigure}{0.32\textwidth}
    \centering
    \includegraphics[width=\textwidth]{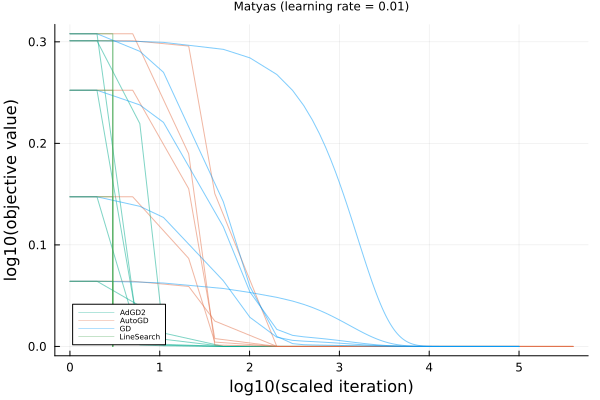}
  \end{subfigure} 
  \begin{subfigure}{0.32\textwidth}
    \centering
    \includegraphics[width=\textwidth]{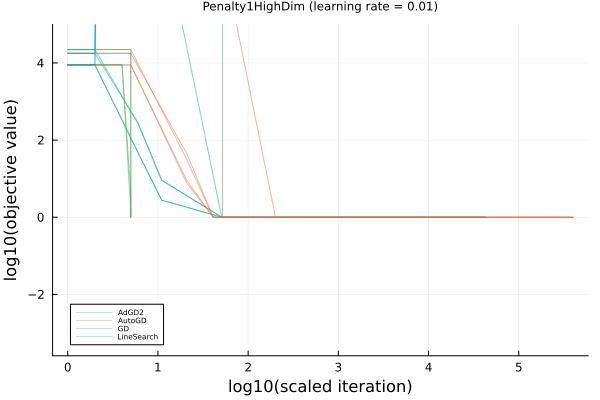}
  \end{subfigure}
  \begin{subfigure}{0.32\textwidth}
    \centering
    \includegraphics[width=\textwidth]{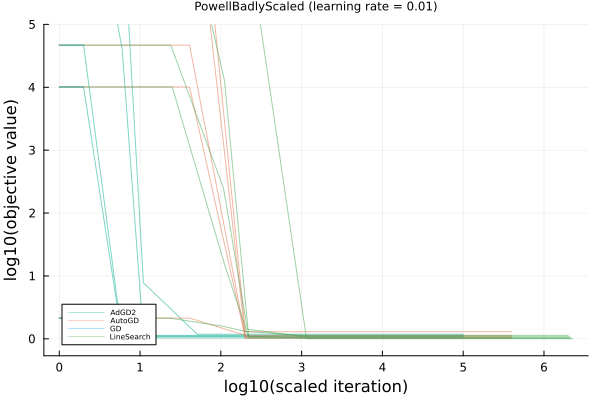}
  \end{subfigure} 
  \begin{subfigure}{0.32\textwidth}
    \centering
    \includegraphics[width=\textwidth]{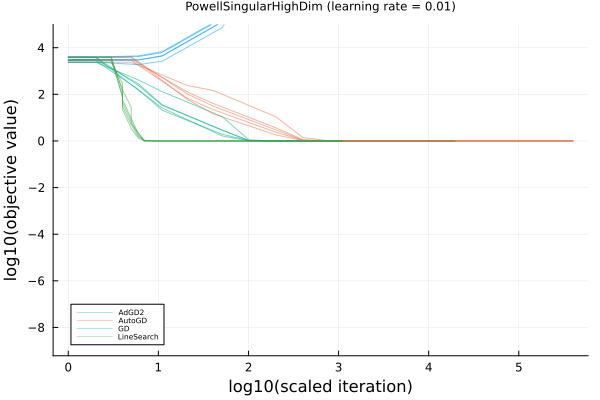}
  \end{subfigure}
  \begin{subfigure}{0.32\textwidth}
    \centering
    \includegraphics[width=\textwidth]{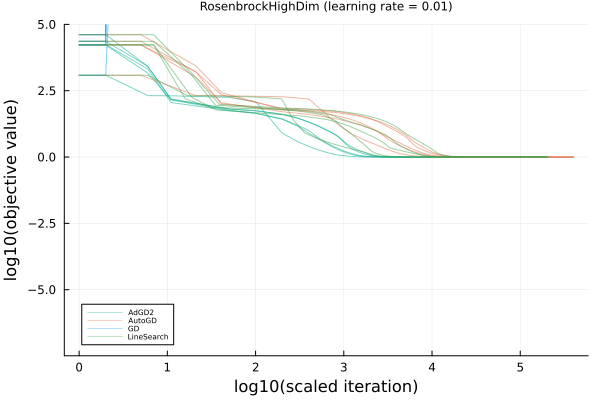}
  \end{subfigure}
  \begin{subfigure}{0.32\textwidth}
    \centering
    \includegraphics[width=\textwidth]{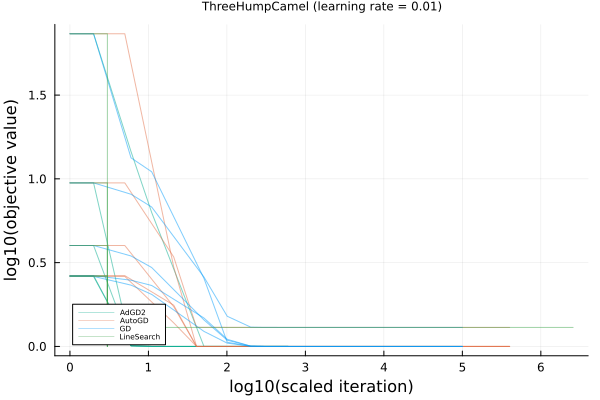}
  \end{subfigure} 
  \begin{subfigure}{0.32\textwidth}
    \centering
    \includegraphics[width=\textwidth]{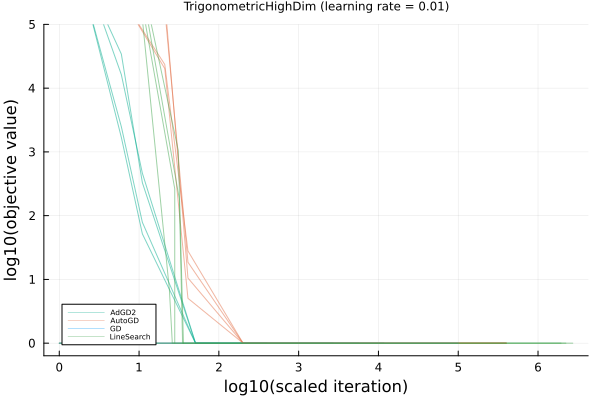}
  \end{subfigure}
  \begin{subfigure}{0.32\textwidth}
    \centering
    \includegraphics[width=\textwidth]{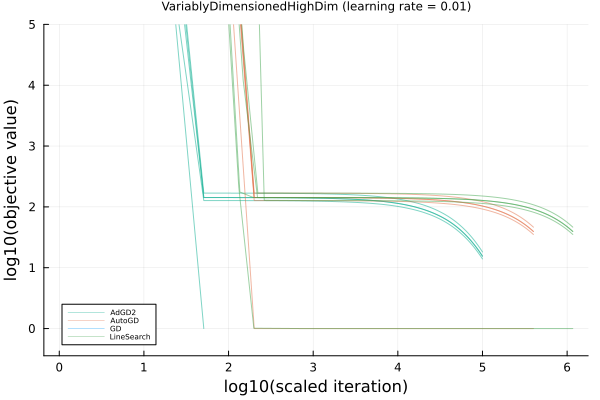}
  \end{subfigure} 
  \begin{subfigure}{0.32\textwidth}
    \centering
    \includegraphics[width=\textwidth]{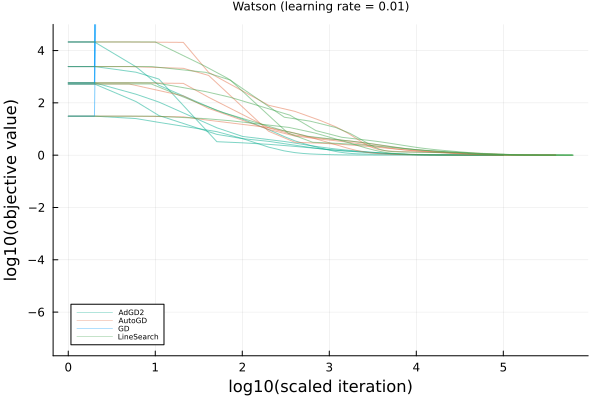}
  \end{subfigure}
  \begin{subfigure}{0.32\textwidth}
    \centering
    \includegraphics[width=\textwidth]{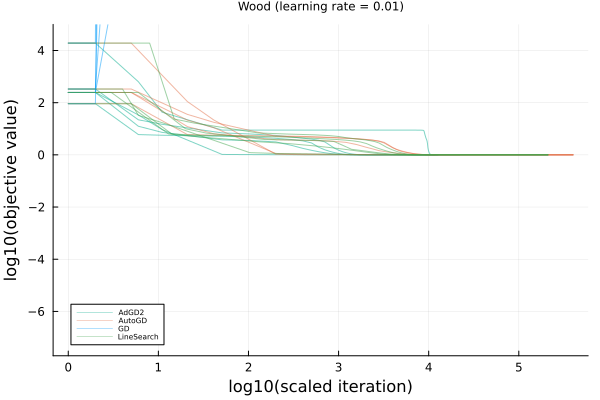}
  \end{subfigure}
  \caption{A subset of the classical optimization experiments with $\gamma_0 = 0.01$ (or initialized with 
  a diffuse distribution concentrated nearby).}
  \label{fig:additional_classical_experiments}
\end{figure*}

\begin{figure*}[!t]
  \centering
  \begin{subfigure}{0.32\textwidth}
    \centering
    \includegraphics[width=\textwidth]{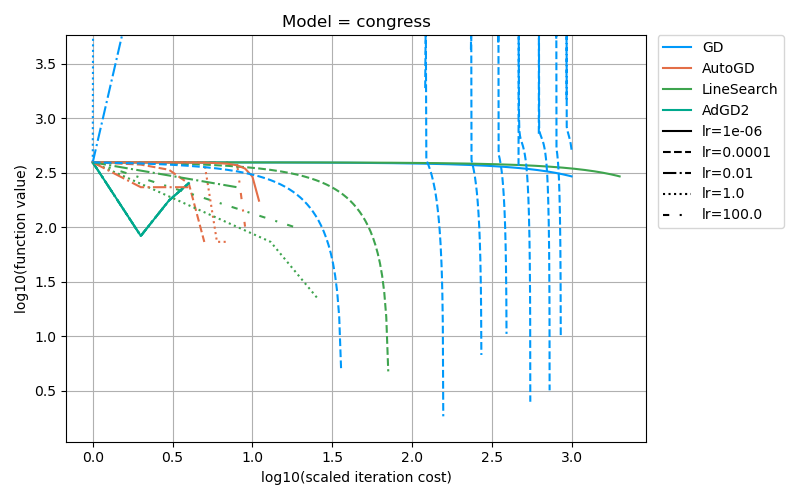}
  \end{subfigure} 
  \begin{subfigure}{0.32\textwidth}
    \centering
    \includegraphics[width=\textwidth]{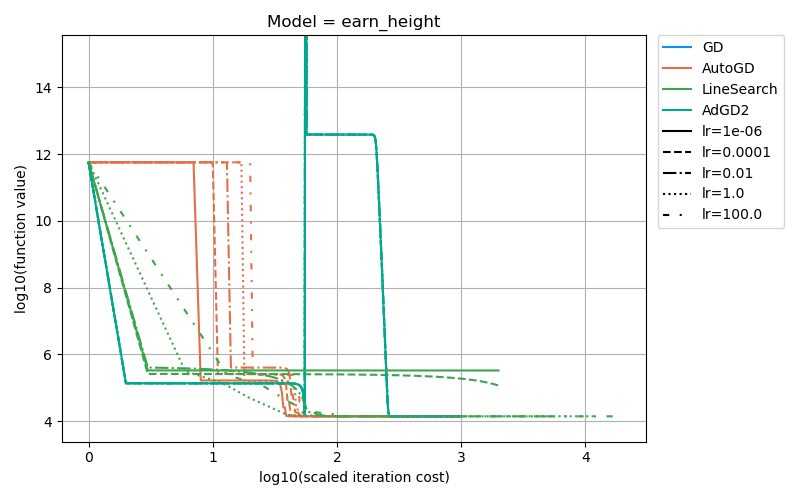}
  \end{subfigure}
  \begin{subfigure}{0.32\textwidth}
    \centering
    \includegraphics[width=\textwidth]{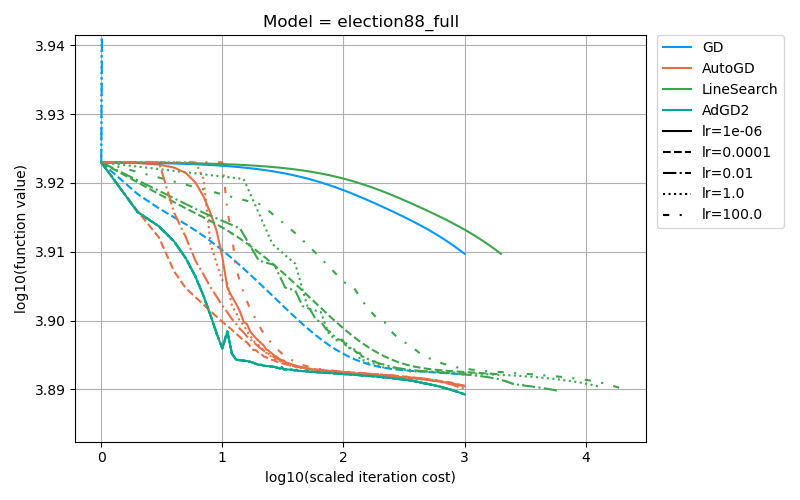}
  \end{subfigure}
  \begin{subfigure}{0.32\textwidth}
    \centering
    \includegraphics[width=\textwidth]{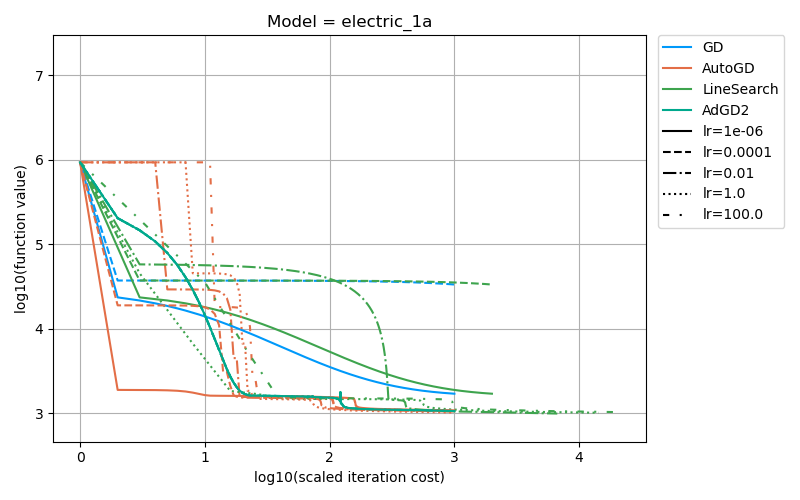}
  \end{subfigure} 
  \begin{subfigure}{0.32\textwidth}
    \centering
    \includegraphics[width=\textwidth]{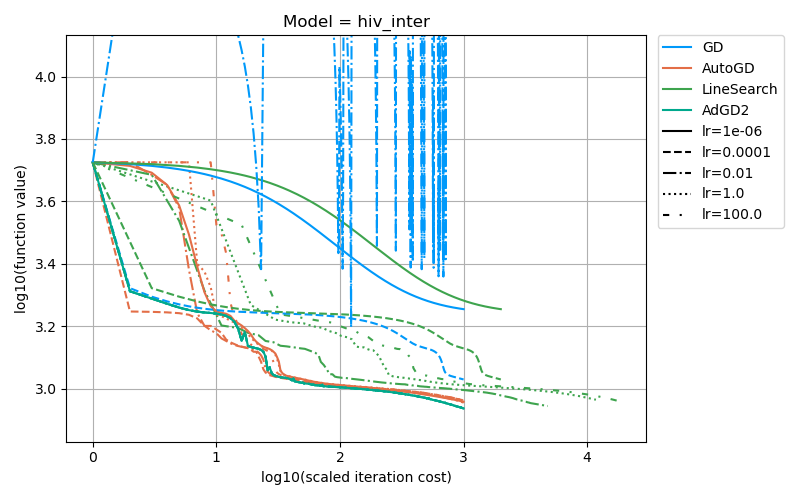}
  \end{subfigure}
  \begin{subfigure}{0.32\textwidth}
    \centering
    \includegraphics[width=\textwidth]{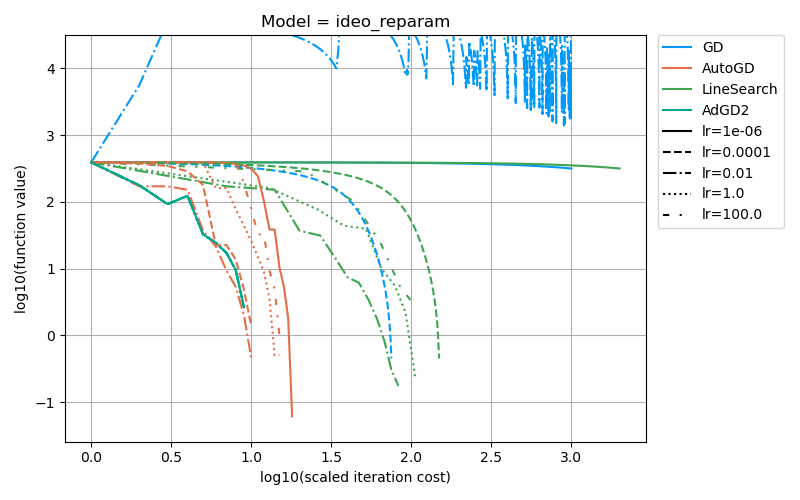}
  \end{subfigure}
  \begin{subfigure}{0.32\textwidth}
    \centering
    \includegraphics[width=\textwidth]{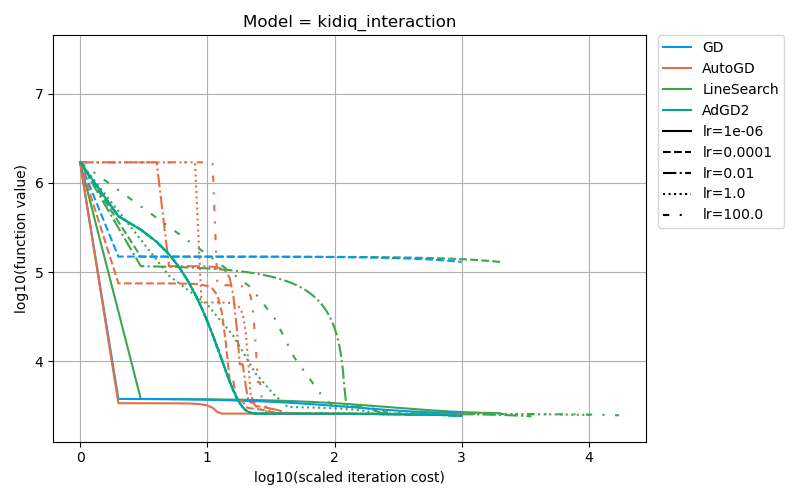}
  \end{subfigure} 
  \begin{subfigure}{0.32\textwidth}
    \centering
    \includegraphics[width=\textwidth]{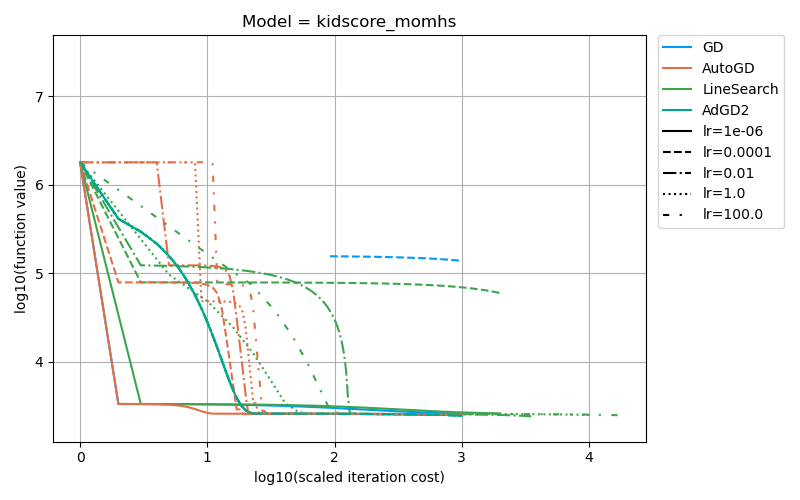}
  \end{subfigure}
  \begin{subfigure}{0.32\textwidth}
    \centering
    \includegraphics[width=\textwidth]{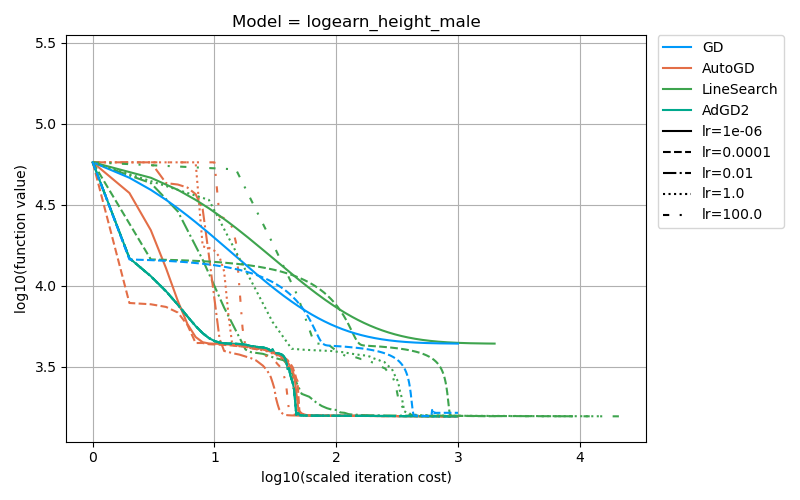}
  \end{subfigure}
  \begin{subfigure}{0.32\textwidth}
    \centering
    \includegraphics[width=\textwidth]{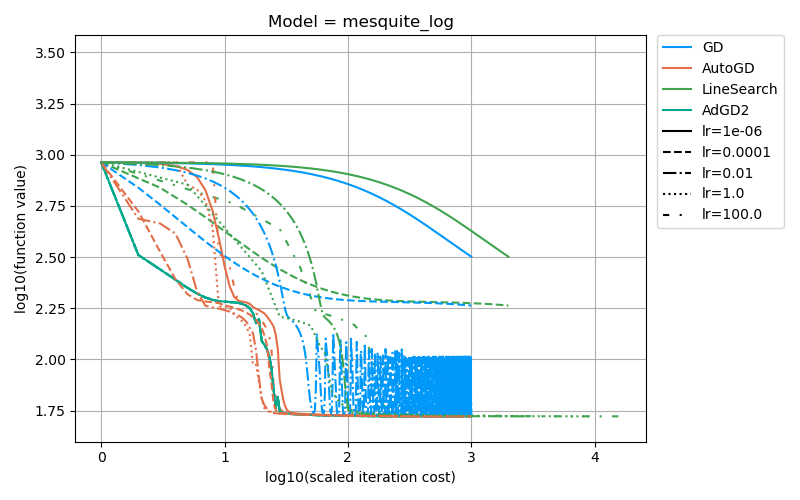}
  \end{subfigure} 
  \begin{subfigure}{0.32\textwidth}
    \centering
    \includegraphics[width=\textwidth]{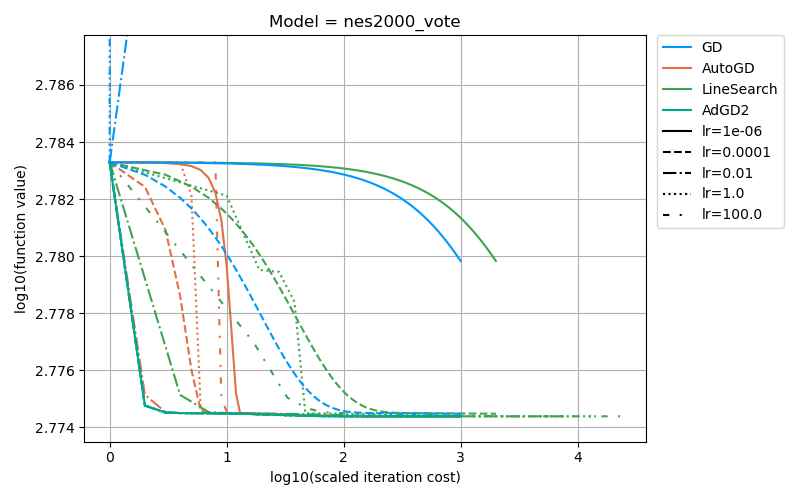}
  \end{subfigure}
  \begin{subfigure}{0.32\textwidth}
    \centering
    \includegraphics[width=\textwidth]{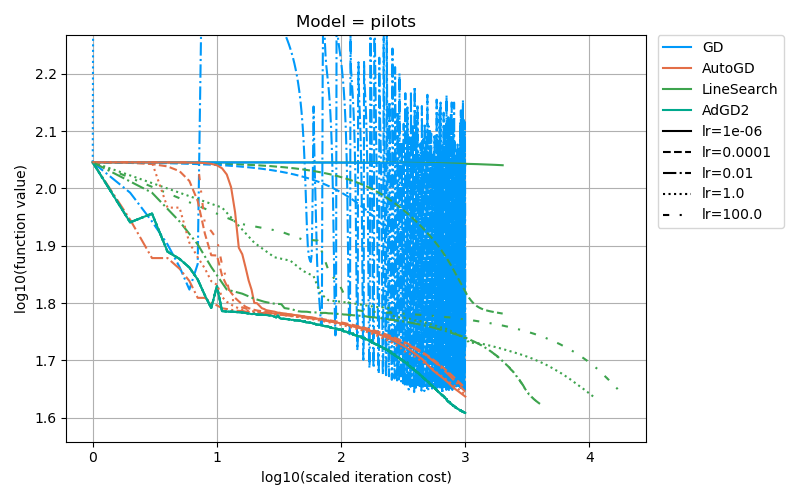}
  \end{subfigure}
  \begin{subfigure}{0.32\textwidth}
    \centering
    \includegraphics[width=\textwidth]{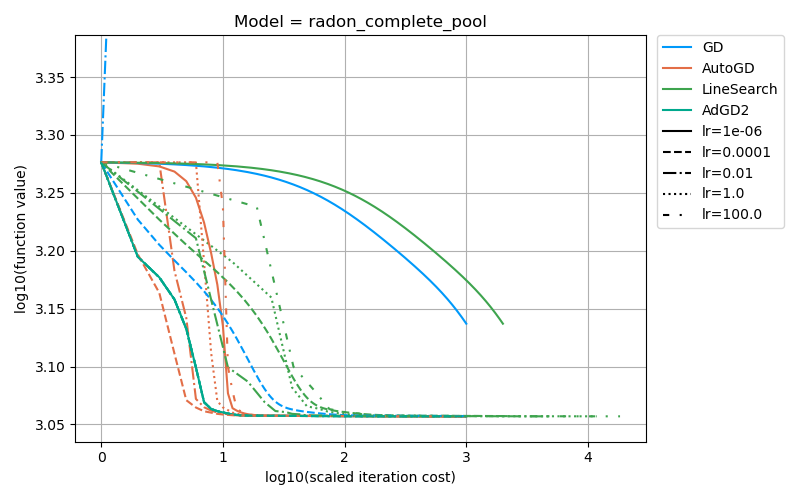}
  \end{subfigure} 
  \begin{subfigure}{0.32\textwidth}
    \centering
    \includegraphics[width=\textwidth]{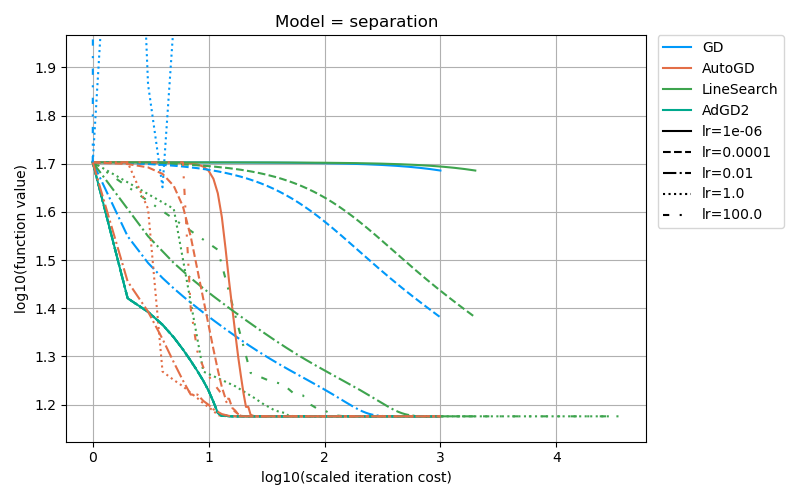}
  \end{subfigure}
  \begin{subfigure}{0.32\textwidth}
    \centering
    \includegraphics[width=\textwidth]{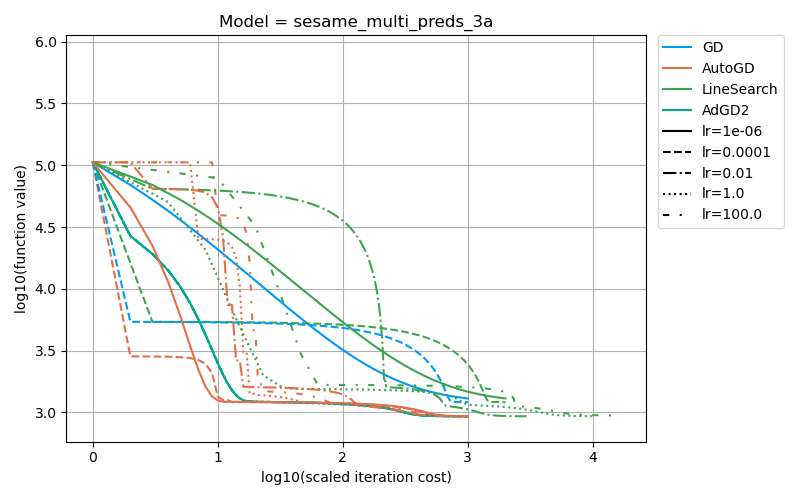}
  \end{subfigure}
  \begin{subfigure}{0.32\textwidth}
    \centering
    \includegraphics[width=\textwidth]{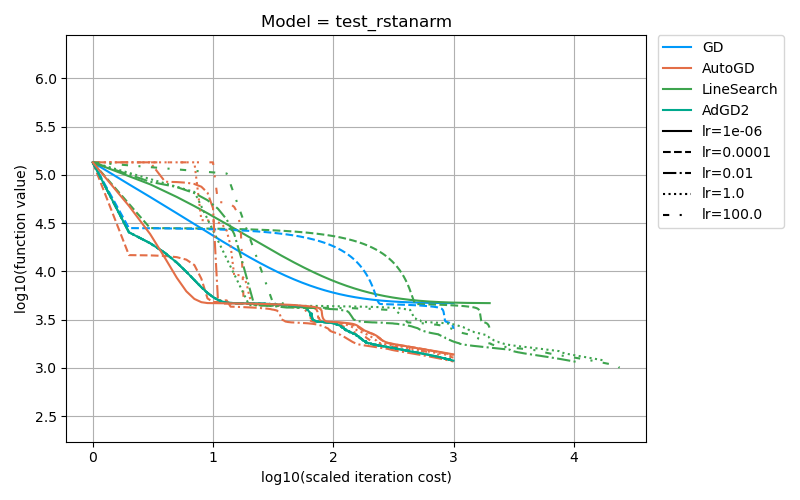}
  \end{subfigure} 
  \begin{subfigure}{0.32\textwidth}
    \centering
    \includegraphics[width=\textwidth]{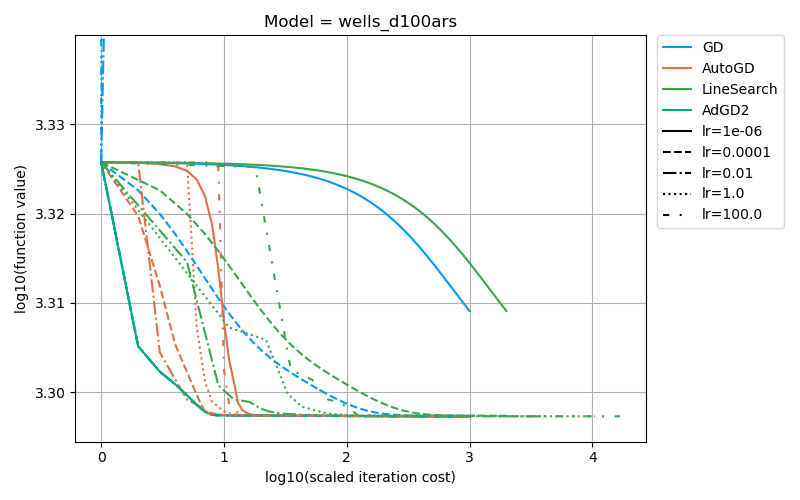}
  \end{subfigure}
  \begin{subfigure}{0.32\textwidth}
    \centering
    \includegraphics[width=\textwidth]{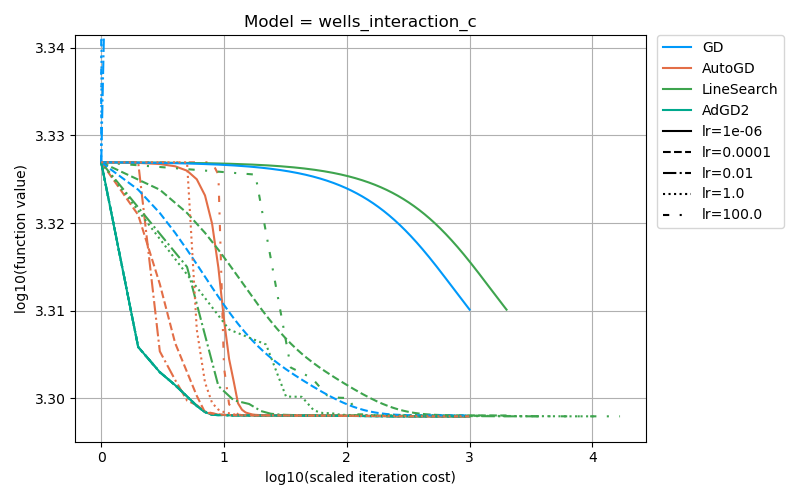}
  \end{subfigure}
  \caption{A subset of the 61 variational inference experiments.}
  \label{fig:additional_dadvi_experiments}
\end{figure*}

Additional results for the classical optimization experiments and the 
variational inference experiments are presented in 
\cref{fig:additional_classical_experiments} and 
\cref{fig:additional_dadvi_experiments}, respectively. 
These figures show the values of the objective functions as each optimizer progresses 
through training. 
Because of the large number of simulation settings considered (87 optimization problems and 
4 optimizers, each with several seeds and up to 100,000 training iterations), 
not all optimization problems are presented in these figures. 
The success curves of \cref{fig:classical_success_curves,fig:dadvi_success_curves} 
better capture aggregate performance across all simulation runs.

\subsection{Pseudocode for AutoBFGS and AutoLBFGS}
\label{sec:autobfgs}

The pseudocode for the AutoBFGS and AutoLBFGS algorithms are provided in 
\cref{alg:autobfgs} and \cref{alg:autolbfgs}, respectively.

\begin{algorithm}[t]
\caption{AutoBFGS}
\begin{algorithmic}[1]
\Require Diffuse initial point $(x_0, \gamma_0) \in \reals^d \times \reals_{>0}$, 
\Require Initial Hessian approximation $H$ (default: $H = I_d$), 
\Require Learning rate scaling coefficient $c$ (default $c=2$), 
\Require Armijo rule constant $\eta$ (default $\eta = 10^{-4}$),
\Require Number of iterations $T$.
\For{$t$ {\bf in} $\{0, 1, \ldots, T\}$}
  \State $p \gets -H \cdot g$ \Comment{preconditioned gradient}
  \State $x_{t+1}, \gamma_{t+1}, \gamma_{t+1}' \gets \texttt{AutoGDStep}(x_t, \gamma_t, -p, c, \eta)$
    \Comment{use AutoGD in this direction}
  \If{$\gamma_{t+1}' > 0$}
      \State $s \gets \gamma_{t+1}' \cdot p$
      \State $y \gets \nabla f(x_{t+1}) - \nabla f(x_t)$
      \State $\rho \gets y^\top s$
      \If{$\rho > 10^{-12}$}
          \State $H \gets (I - \frac{s y^\top}{\rho}) H (I - \frac{y s^\top}{\rho}) + \frac{s s^\top}{\rho}$
      \EndIf
  \EndIf
\EndFor
\label{alg:autobfgs}
\end{algorithmic}
\end{algorithm}

\begin{algorithm}[t]
\caption{AutoLBFGS}
\begin{algorithmic}[1]
\Require Diffuse initial point $(x_0, \gamma_0) \in \reals^d \times \reals_{>0}$, 
\Require Memory size $m$ (default $m=10$), 
\Require Learning rate scaling coefficient $c$ (default $c=2$), 
\Require Armijo rule constant $\eta$ (default $\eta = 10^{-4}$),
\Require Number of iterations $T$.
\State $S, Y, P \gets [\,], [\,], [\,]$ 
  \Comment {initialize empty histories for $s$, $y$, and $\rho$}
\For{$t$ {\bf in} $\{0, 1, \ldots, T\}$}
  \State $p \gets \texttt{LBFGSDirection}(S, Y, P, \nabla f(x_t))$
    \Comment{preconditioned gradient using two-loop recursion}
  \State $x_{t+1}, \gamma_{t+1}, \gamma_{t+1}' \gets \texttt{AutoGDStep}(x_t, \gamma_t, -p, c, \eta)$
    \Comment{use AutoGD in this direction}
  \If{$\gamma_{t+1}' > 0$}
      \State $s \gets \gamma_{t+1}' \cdot p$
      \State $y \gets \nabla f(x_{t+1}) - \nabla f(x_t)$
      \State $\rho \gets y^\top s$
      \If{$\rho > 10^{-12}$}
          \State Append $s$ to $S$, $y$ to $Y$, and $1/\rho$ to $P$
          \If{$|S| > m$} 
              \State Remove oldest entries from $S$, $Y$, and $P$
          \EndIf
      \EndIf
  \EndIf
\EndFor
\label{alg:autolbfgs}
\end{algorithmic}
\end{algorithm}

\end{document}